\documentclass{article}
\usepackage{imports/arxiv}
\usepackage{amssymb}
\usepackage{amsmath}
\usepackage{graphicx}
\usepackage[mathscr]{euscript}
\usepackage{lineno}
\usepackage{subcaption}
\usepackage[utf8]{inputenc}
\usepackage{float}
\usepackage[normalem]{ulem}
\usepackage{multicol}
\usepackage{lmodern}
\usepackage{lipsum}
\usepackage{marvosym}
\usepackage{mathtools}
\usepackage{soul}
\usepackage{bm}
\usepackage{amsthm}
\usepackage{xargs}                      
\usepackage{xcolor}
\usepackage{bbm}
\usepackage{hyperref}
\usepackage[title]{appendix}

\usepackage{svg}
\usepackage{xcolor}
\usepackage{mathtools}

\makeatletter
\def\mathcolor#1#{\@mathcolor{#1}}
\def\@mathcolor#1#2#3{%
  \protect\leavevmode
  \begingroup
    \color#1{#2}#3%
  \endgroup
}
\makeatother

\renewcommand{\eqref}[1]{{Eq.~\ref{#1}}}

\newcommand\restr[2]{{
  \left.\kern-\nulldelimiterspace 
  #1 
  \vphantom{
  \big|
  } 
  \right|_{#2} 
  }}
  
\makeatletter
\@ifclassloaded{beamer}{}{

    \theoremstyle{definition}
    \newtheorem{definition}{Definition}

    \theoremstyle{remark}
    
}
\makeatother

\makeatletter
\renewcommand{\boxed}[1]{\text{\fboxsep=.2em\fbox{\m@th$\displaystyle#1$}}}
\makeatother

\definecolor{orange}{RGB}{255,127,0}
\definecolor{blackchocolate}{HTML}{191102}
\definecolor{rosewood}{HTML}{510d0a}
\definecolor{rufous}{HTML}{A31B14}
\definecolor{citron}{HTML}{a29f15}
\definecolor{orangeyellow}{HTML}{f3b61f}
\definecolor{teagreen}{HTML}{bbd8b3}
\definecolor{gray}{RGB}{128,128,128}
\definecolor{lightcitron}{RGB}{189,187,91}
\definecolor{lightrufous}{RGB}{190,95,90}
\definecolor{steelblue}{HTML}{4C86A8}

\usepackage{imports/fonts}
\usepackage[numbers]{natbib}
\usepackage{authblk}

\title{Deep Learning for Forensic Identification of Source}
\author[1]{Cole Patten}
\author[1]{Christopher Saunders}
\author[1]{Michael Puthawala}
\affil[1]{Department of Mathematics \& Statistics, South Dakota State University}
\affil[ ]{\textit {\{cole.patten, christopher.saunders, michael.puthawala\}@sdstate.edu}}

\date{\today}

\begin{document}

\maketitle

\begin{abstract}

    We used contrastive neural networks to learn useful similarity scores between the 144 cartridge casings in the NBIDE dataset \citep{vorburger2007surface}, under the common-but-unknown source paradigm. The common-but-unknown source problem is a problem archetype in forensics where the question is whether two objects share a common source (e.g. were two cartridge casings fired from the same firearm). Similarity scores are often used to interpret evidence under this paradigm. We directly compared our results to a state-of-the-art algorithm, Congruent Matching Cells (CMC) \citep{song2015proposed, tong2015improved}. When trained on the E3 dataset \citep{basu2022forensic} of 2967 cartridge casings, contrastive learning achieved an ROC AUC of 0.892. The CMC algorithm achieved 0.867. We also conducted an ablation study where we varied the neural network architecture; specifically, the network's width or depth. The ablation study showed that contrastive network performance results are somewhat robust to the network architecture. This work was in part motivated by the use of similarity scores attained via contrastive learning for standard evidence interpretation methods such as score-based likelihood ratios. 

\end{abstract}

\section{Introduction}

\subsection{Background}

If two cartridge casings were found on a crime scene, a natural question is ``were these cartridges fired from a single firearm or multiple firearms?'' This is an example of a \emph{common-but-unknown source problem} \citep{ommen2018building}, where it must be determined if two objects are from the same source or different sources. This problem cannot be properly addressed under a classification paradigm.

 The example of cartridge casings found at a crime scene makes the issue clear. Under a classification paradigm, you begin with a list of $k$ classes (in this case the classes are firearms) called the gallery. Each cartridge casing is assigned to one class from the gallery. If they are assigned to the same firearm you say the cartridges were fired from the same firearm. This method does not account for the possibility that one or both of the cartridges were fired by a firearm that is not in the gallery. Using the classification paradigm for common-but-unknown source problems gives rise to a type I error if two cartridge casings are both assigned to the same firearm from the gallery, but one or both were fired from firearms not from the gallery. Similarly, this paradigm creates the potential for a type II error if the cartridge casings are assigned to two different firearms from the gallery, but were fired from the same firearm which was not in the gallery. 

Prior works applying deep learning to cartridge casing comparison fail to simultaneously consider the common-but-unknown source problem and the possibility of a non-exhaustive gallery of sources. This work uses deep learning for this task. 

\begin{figure}[htpb]
      \centering
      \includegraphics[width=.6\textwidth]{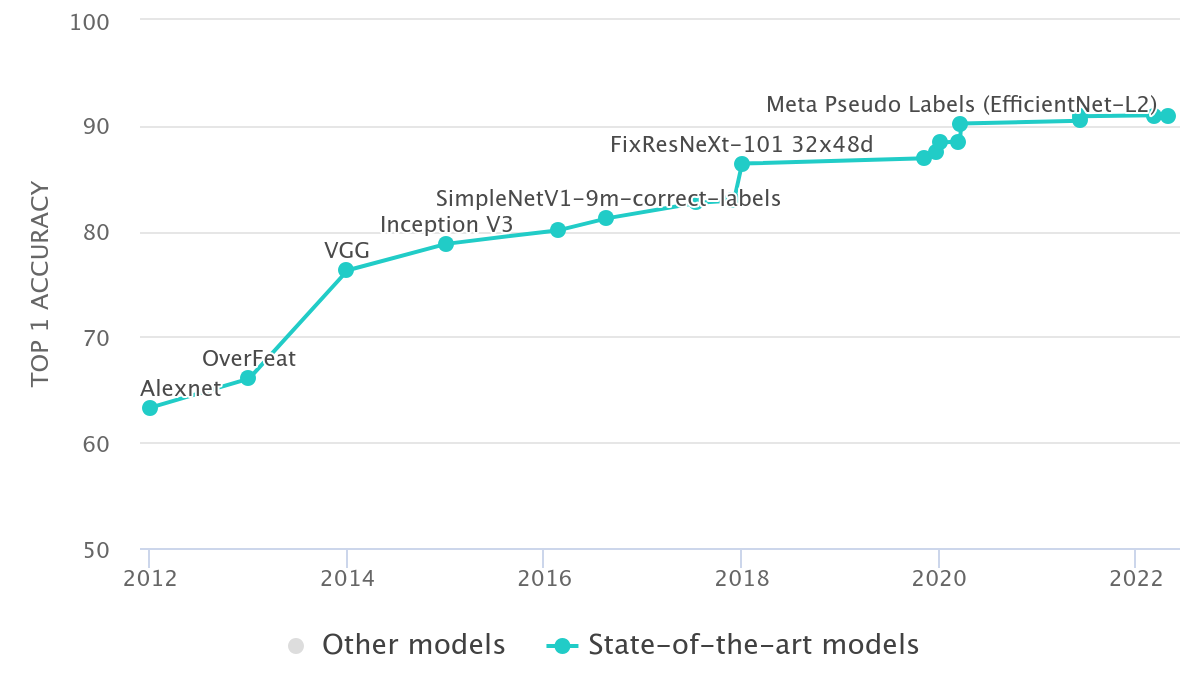}
      \caption{Graph depicting state-of-the-art classification accuracy of neural networks on the ImageNet database. (Source: \citet{paperswithcode})}
      \label{fig:imagenet}
\end{figure}

The field of deep learning --in general-- has rapidly advanced various fields, from computer vision to language processing. This progress is well exemplified by the ever-increasing capabilities of neural networks to correctly classify images from the ImageNet database \citep{deng2009imagenet}. Figure \ref{fig:imagenet} shows how neural networks, in just a decade, have gone from $63.3\%$ \citep{krizhevsky2012imagenet} to $78.3\%$ accuracy \cite{yu2022coca}. Given the success of neural networks on image classification problems, it is reasonable to expect that they may perform well in forensic identification of source problems. In this work, we do precisely this.

\subsection{Prior Work}

\subsubsection{Deep Learning, Contrastive Learning}

Deep learning has previously been used to analyze cartridge casings scans \citep{kamaruddin2011firearm, Kamaruddin2012Firearm, giudice2019siamese, valentim2021gun}. To suitably deploy neural networks on real-world forensic identification or source problems, it must be first demonstrated that they can effectively identify the similarities between cartridge casings fired from the same \textit{specific} firearm and contrast casings fired from \textit{distinct} firearms. Furthermore, these abilities should generalize to firearms not present in the network's training dataset.

Contrastive learning was introduced in \citet{siamese}, by learning (dis)similarity scores. As early as 2011, \citet{kamaruddin2011firearm} demonstrated the ability of neural networks to accurately classify the firearm that fired a cartridge casing. In \citet{giudice2019siamese} it was shown that contrastive neural networks are capable of discriminating between cartridge casings fired by different firearms, but only when all firearms are present in the training dataset. In other words, when the training gallery is exhaustive. In \citet{valentim2021gun}, contrastive networks were shown to be able to discriminate between cartridge casings fired by different \textit{models of firearms}, but not necessarily between two firearms of the same model. More recently, contrastive networks were applied to images of casings in \citet{patten2024contrastive} to verify if two casings were fired by the same \textit{specific firearm}, but the proposed method failed to beat the state-of-the-art method that didn't use neural networks.

\subsubsection{Forensics}

There has been a recent trend to create objective, computer-based methods to aid human examiners and decrease subjectivity in identification of source problems pertaining to cartridge casings \citep{mattijssen2023interpol}. 

A popular \citep{mattijssen2023interpol} method for assessing the similarity of cartridge casings is the Congruent Matching Cells (CMC) algorithm \citep{song2015proposed}. Since its inception, the CMC algorithm has been the subject of multiple papers improving the algorithm \citep{tong2015improved, chen2017convergence} and adjusting the algorithm to firing pin impressions \citep{zhang2016correlation}. Because there is no learning required for CMC (i.e. there is no optimization required) the algorithm has some advantage over deep learning methods. 

Other works utilize feature extraction methods such as geometric moments of cartridge casing images. For example, \citet{basu2022forensic} uses linear discriminant functions on the extracted feature vectors to calculate likelihood ratios. Using geometric moments rather than whole images as inputs for neural networks, \citet{Kamaruddin2012Firearm} found an improvement in classification accuracy.

\subsection{Our Contribution}

Our contributions are twofold.

First, we use contrastive neural networks on cartridge casings to address the common-but-unknown source problem. That is, we use neural networks to provide an answer to the question ``Were these cartridge casings fired from the same firearm?'' Further, we restrict ourselves to the common-but-unknown source problem for cartridges fired from guns that were not included in the training dataset. We believe this better models how these methods would be used in the field, where the training gallery cannot be assumed to be exhaustive. We obtained a neural network architecture that achieves a higher ROC AUC (Definition \ref{def:ROCAUC} on the NBIDE dataset than the CMC algorithm. The ability of contrastive neural networks to compete with modern forensic statistical techniques suggests that they may be useful for practitioners in the interpretation of evidence. 

Secondly, we conduct an ablation study where we modify our neural network's width and depth. We compare the effectiveness of our networks by their ROC AUC on the NBIDE dataset. In this way we determine an optimal network architecture for our given training dataset, and measure the network's sensitivity to architectural changes. 

Our code is available at \url{https://github.com/colepatten/DeepLearningSourceID/}.

\section{Methods}

\subsection{Datasets}

\begin{figure}[htpb]
      \centering
      \includegraphics[width=.25\textwidth]{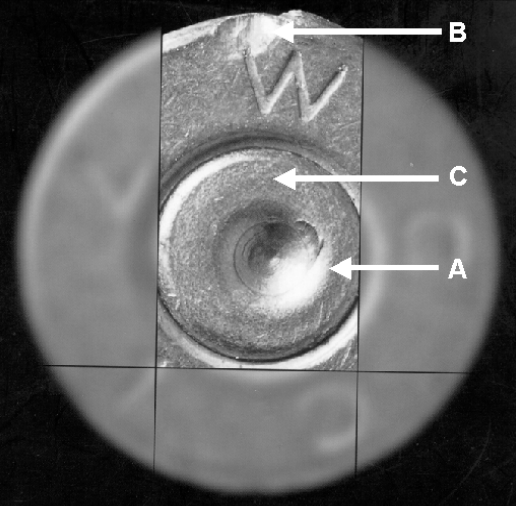}
      \caption{Image showing the A) firing pin impression, B) ejector mark, and C) breech face impression. Although all regions of the cartridge casing may be utilized as forensic evidence, the CMC algorithm is designed to operate exclusively on the breech face. Therefore, the breech face of the cartridge casing is the only region considered in this work. (Source: \citet{vorburger2007surface})}
      \label{fig:impressions}
\end{figure}

In this study, we considered 3D topography scans of 9mm cartridge casings. An example of a 9mm cartridge casing is shown in Figure \ref{fig:impressions}. To maintain parity with the CMC algorithm, we considered only the breech face region of the cartridge casings. Two datasets were used, both were downloaded from and are hosted on \citet{NBTD}.

The NBIDE dataset \citep{vorburger2007surface} contains 144 cartridges. The cartridges were fired from 12 firearms. There are three distinct firearm models, and four of each model which comprise the 12 firearms. Each firearm in the NBIDE dataset fired 12 cartridges. The cartridges are from four different manufacturers.

The E3 dataset \citep{basu2022forensic} contains 2969 cartridges; however, only 2967 were used in this study. The cartridges were fired from 297 firearms of varying brands and models. Each firearm in the dataset fired 9 or 10 cartridges. The cartridges themselves came from a variety of manufacturers. 

\subsection{Contrastive Learning}
We implemented contrastive learning to train a neural network for verification. We used residual networks \citep{he2016deep} as our network architecture (see Figure \ref{fig:backref}) and opted for supervised contrastive loss \citep{khosla2020supervised} as our loss function. 

\subsubsection{Architecture}

\begin{figure}[htpb]
      \centering
      \includegraphics[width=.6\textwidth]{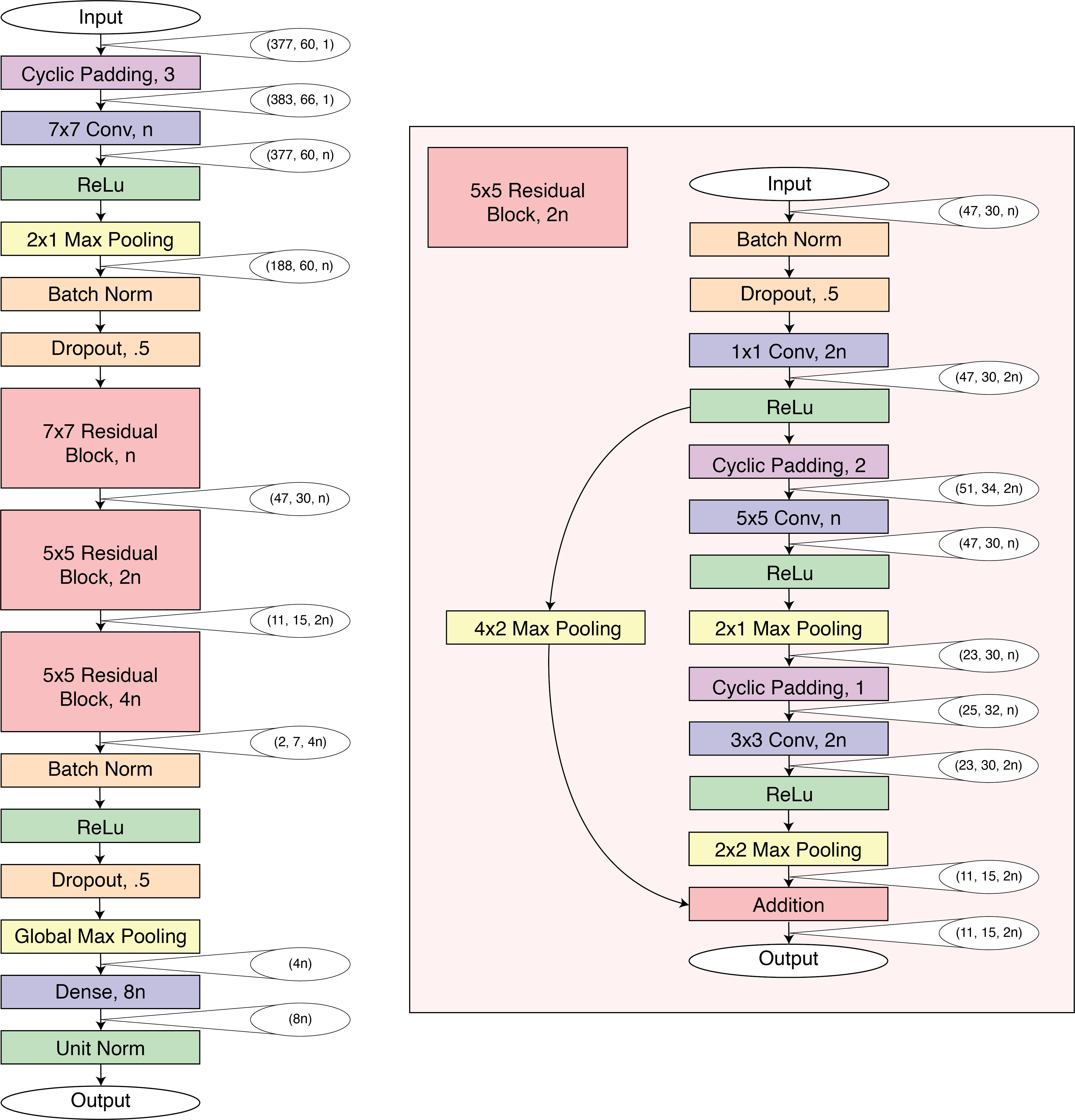}
      \caption{(Left) The shared architecture of all neural networks used in the study. Models varied in the composition of their residual blocks and specific choices of $n$. (Right) The first 5x5 residual block in our top-performing Reference Model. In both diagrams, the bubbles to the right of the model show the dimensions of a cartridge casing scan as it passes through the network.}
      \label{fig:backref}
\end{figure}

All of the neural networks used in this work were residual neural networks \citep{he2016deep} which allow the training of deeper neural networks \citep{li2018visualizing}. What defines a residual neural network is that every few layers of the network are divided into \emph{residual blocks}. As data, $x$, passes through a residual block, $b(\cdot)$, the input to each block is added to the output of the block in a \emph{skip connection} ($y = b(x) + x$).  

The right side of Figure \ref{fig:backref}, in the pink box, depicts the specific composition of a residual block in our top-performing Reference Model. The long arrows to the left of the main branch of the model are the skip connections. The Double Block, Block Depth=2, and Block Depth=4 networks are different from the Reference Model by variations of the residual block shown here. The specific compositions of their residual blocks are depicted in Figure \ref{fig:other_models}.

On the left side of Figure \ref{fig:backref}, the shared residual network architecture of all our models is shown. For the Width=8, Width=32, and Width=64 networks, the entire architecture is the same as our Reference Model, but with the width parameter $n$ changed accordingly. For the Reference Model, the parameter is set to $n= 16$. 

Details on the exact number of layers and parameters in each model are presented in Figure \ref{tab:info}.

An important aspect of our network architecture is the use of the cyclic padding layers (colored in purple in Figure \ref{fig:backref}) which account for the rotational invariance of cartridge casing images. Rotating an image of a cartridge casing has no effect on which gun fired the cartridge casing. We would like our contrastive network to use this information to its advantage. 

One approach to obtain rotational invariance is through data augmentation, that is, rotating the training images. In this way, the contrastive network embeds an image $x$ and its rotated counterpart $x^*$ close together.

We found it to be more effective to convert our images to polar coordinates and add cyclic padding along the angular dimension before each convolutional layer. 

To be concrete, let $x\in \mathbb{R}^{60\times377}$ be our image in polar coordinates where the first dimension corresponds to the angle to the positive x-axis and the second corresponds to the radius. We can write $x$ in terms of its column vectors as $x=[x^1,x^2,\dots,x^{377}]$ where each $x^i\in\mathbb{R}^{60}$. Then, before a convolutional layer with a kernel $k\in\mathbb{R}^{5\times5}$, we will add cyclic padding to the image to obtain 
\begin{align}
    x^*\coloneqq [x_{376}, x_{377}, x_{1}, x_2, \dots,x_{377}, x_1, x_2].
\end{align}

This method incurs some error from interpolating the values from the original image to obtain the image in polar coordinates. With this method, however, there is no need to optimize for rotational invariance, because it is already accounted for.

\subsubsection{Loss Function}

\begin{figure}[htpb]
      \centering
      \includegraphics[width=.6\textwidth]{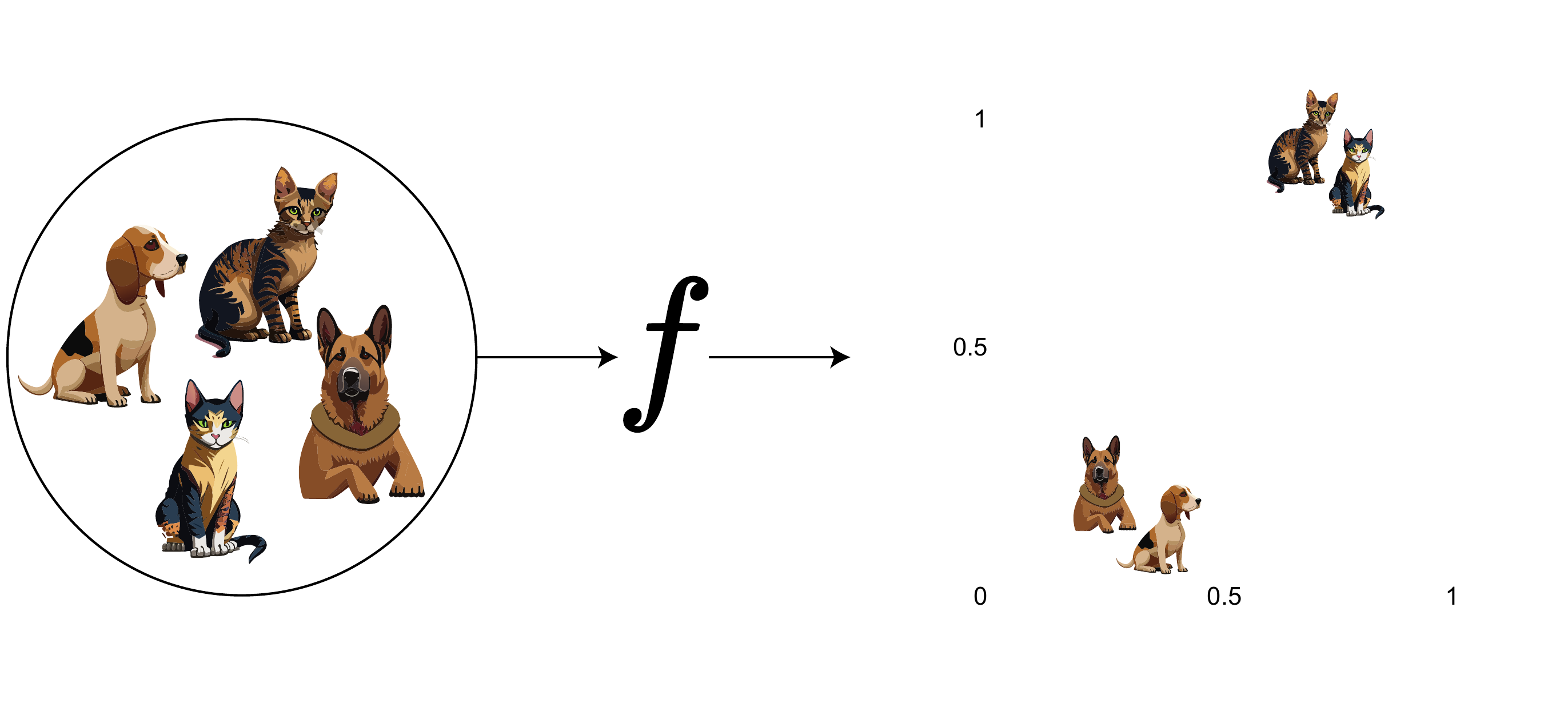}
      \caption{Raw data may exist strewn about in its ambient data space. For example, 224x224 images can be thought of as existing in $\mathbb{R}^{224\times224}$. In this space, images and dogs and cats may not be easily separated. The goal of contrastive learning is to learn an embedding function, $f$, that maps the data from its chaotic ambient space to an embedding space where there is more order. In the dogs in cats examples, the increased order we want is for images of dogs and cats to exist in their own distinct clusters.}
      \label{fig:contrastive}
\end{figure}

Contrastive neural networks are optimized to learn an embedding function that maps objects from the same class close together, and objects from different classes distant from one another (see Figure \ref{fig:contrastive}). In this work, we trained networks with the supervised contrastive loss function \citep{khosla2020supervised}. Let $S = \{x_1, \dots, x_n\}$ be the sample and $f$ denote the neural network. 

\begin{definition}[SupCon Loss]
    \begin{equation}
        \mathcal{L}_{supcon}(S, f) = \sum_{x_a \in S}\frac{-1}{\left| P(x_a) \right|}\sum_{x_p \in P(x_a)} \log \frac{\exp(f(x_a) \cdot f(x_p) / \tau)}{\sum_{x_n \in N(x_a)} \exp(f(x_a) \cdot f(x_n)) / \tau)}
    \end{equation}
    where $\tau \in \mathbb{R}^+$ is a hyperparameter called the temperature, $P(x_a) \subset S$ denotes the objects in the sample agreeing in class with $x_a$, and  $N(x_a) \subset S$ likewise denotes the objects in the sample differing in class from $x_a$.
\end{definition}

An elucidatory rewriting of the supervised contrastive loss function is 
\begin{align} 
        \mathcal{L}_{supcon}(S, f) &= \sum_{x_a \in S} \left[ \log \left( \sum_{x_n \in N(x_a)} \exp \left( \frac{f(x_a) \cdot f(x_n)}{\tau} \right) \right) - \frac{1}{\left| P(x_a) \right|}\sum_{x_p \in P(x_a)} \frac{f(x_a) \cdot f(x_n)}{\tau} \right] \label{SC-PC4}.
\end{align}
In this form, for each $x_a \in S$ the loss can be viewed as the difference between two terms. The first term is the LogSumExp of $\frac{f(x_a)\cdot f(x_n)}{\tau}$, the dot products divided by $\tau$, for all objects $x_n$ of a different class than $x_a$. The second term is the average modified cosine similarities between $f(x_a)$ and $f(x_p)$ for objects $x_n$ of the same class as $x_a$. Notice that the loss function is decreased by reducing the distance between $f(x_1)$ and $f(x_2)$ where $x_1$ and $x_2$ belong to the same class and increasing the distance between  $f(x_1)$ and $f(x_3)$ where $x_1$ and $x_3$ belong to different classes. 

\subsection{Congruent Matching Cells (CMC)}

We used the cmcR package \citep{zemmels2023cartridge} in R to implement the Congruent Matching Cells (CMC) \citep{song2015proposed} and High CMC \citep{tong2015improved} algorithms for comparing two cartridge case breech faces. In the traditional implementation, a threshold number of matching cells between two casings, above which the casings are declared a match, and below which a non-match. Because we wanted similarity scores, we instead counted the total number of matching cells, divided by 128 (the maximum number of matching cells) and, used this value as our similarity score. 

Note that the CMC algorithm relies on choosing four hyperparameters, $T_x, T_y, T_\theta, T_{CCF}$, which \citet{song2015proposed} suggests should be experimentally chosen. For our study, we chose $T_x = T_y = 0.4, T_\theta=10, T_{CCF}=0.5$. For more information on these threshold parameters, reference the original CMC algorithm proposal \citep{song2015proposed}. The parameters are also described in \citet{zemmels2023cartridge} along with a table of popular choices for their values.

We used the CMC algorithm to compare all pairs of cartridge casings in the NBIDE dataset. The CMC algorithm was implemented in R, with the cmcR library, as described in \citet{zemmels2023cartridge}.

\section{Experiments}

Our goal is to provide a direct comparison between contrastive learning and congruent matching cells in the task of providing useful similarity scores between two cartridges. The usefulness of the methods was measured by the area under the curve of the receiver operating characteristic, and the dataset used for testing was the NBIDE cartridge casing dataset. To train the neural networks, we used the E3 dataset of 2967 spent cartridge casings. 

\subsection{Preprocessing}

\begin{figure}[htpb]
    \centering
    \hspace*{\fill} 
    \begin{minipage}[t]{0.3\textwidth}
        \centering
        \includegraphics[width=\textwidth]{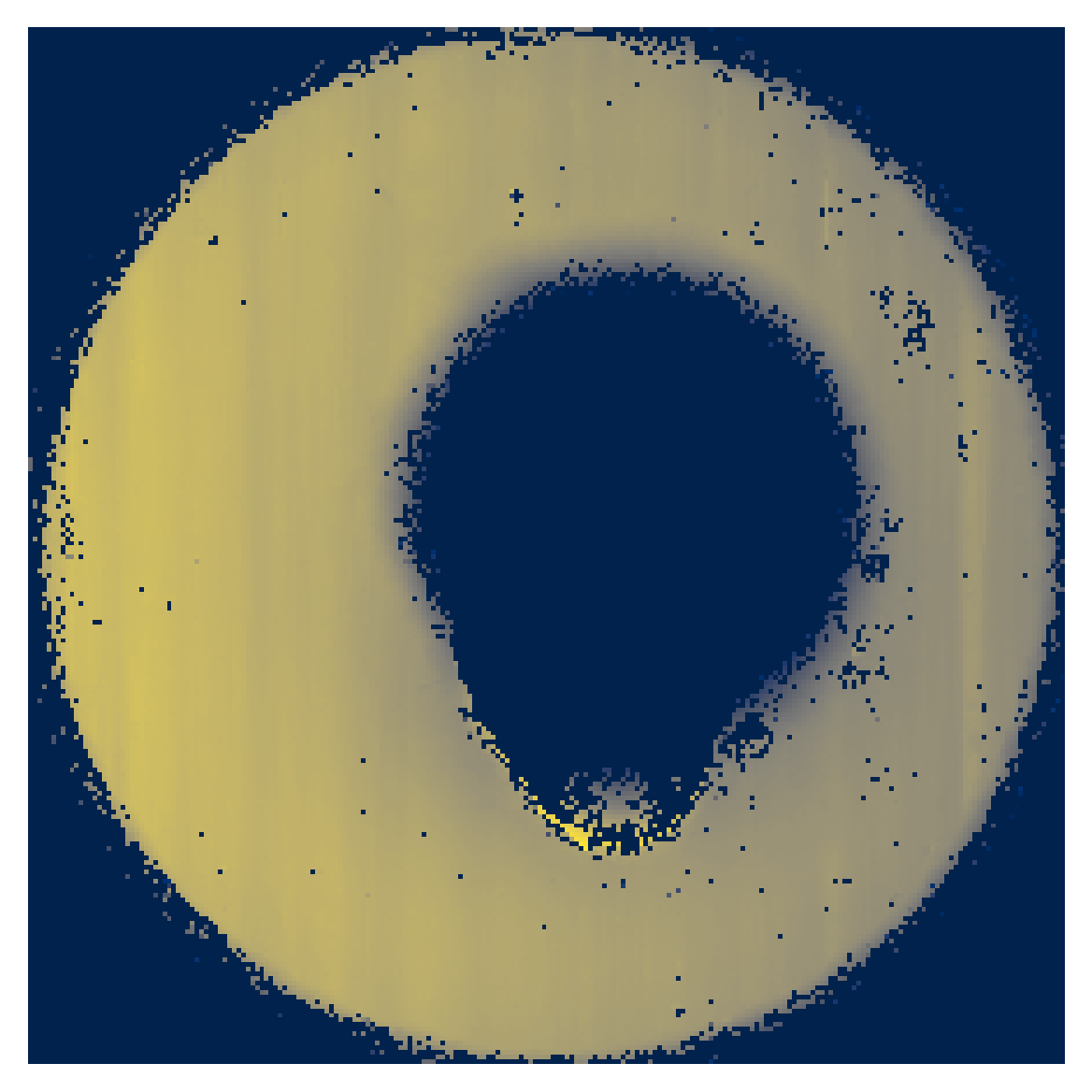}
        \caption*{(a) Raw cartridge casing scan.}
    \end{minipage}
    \hfill
    \begin{minipage}[t]{0.3\textwidth}
        \centering
        \includegraphics[width=\textwidth]{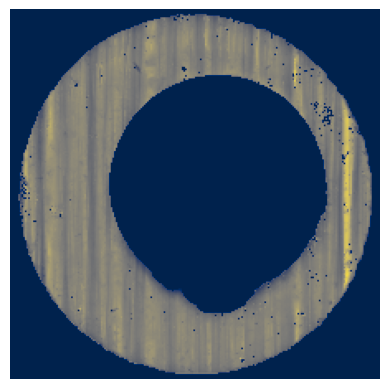}
        \caption*{(b) Cartridge casing scan as preprocessed for use in CMC algorithm.}
    \end{minipage}
    \hspace*{\fill} 

    \vspace{.5em} 
    \begin{minipage}[t]{0.6\textwidth} 
        \centering
        \includegraphics[width=\textwidth]{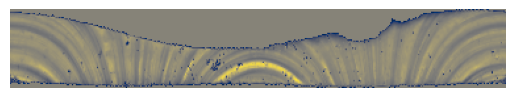}
        \caption*{(c) Cartridge casing as preprocessed for use in contrastive learning.}
    \end{minipage}

    \caption{a) A raw cartridge casing scan from the NBIDE dataset. b) The image of the same cartridge casing after it has gone through the necessary preprocessing steps to be used in CMC. c) The same cartridge casing image from (b) after it has been converted to polar coordinates and centered.}
    \label{fig:image_processing}
\end{figure}

For our implementation of the CMC algorithm, our preprocessing pipeline closely followed that detailed in \citet{zemmels2023cartridge}. Note that the congruent matching cells algorithm compares \textit{only the breech faces} of two spent cartridge casings and does not consider the \textit{firing pin impression} or the \textit{ejection mark}. In Figure \ref{fig:image_processing} (a) the original cartridge casing scan is seen. Figure \ref{fig:image_processing} (b) shows the same scan preprocessed for use in CMC.

For the contrastive networks, the same preprocessing steps are taken as for the CMC algorithm, with a couple of additional steps to represent our images in polar coordinates. Each image was reduced to 224x224 pixels. We then created a 377x60 grid and interpolated the original image to fill in the grid with the correct pixel values in polar coordinates. We also normalized all non-zero pixels in the image to mean 0 and standard deviation 1. Applying this pipeline to the image in Figure \ref{fig:image_processing} (b) results in the in Figure \ref{fig:image_processing} (c).

\subsection{Setup}

We varied the size of our contrastive networks to observe the effects. Our Reference Model, which was our top-performing architecture, is shown in Figure \ref{fig:backref} where $n=16$. In our experiment we varied the 1) network width $n$, 2) the number of residual blocks, and 3) the number of layers within a residual block. We adjusted each of these factors one at a time from their values in the reference model. We compared the methods used in our experiments by their \emph{ receiver operating characteristic area under the curve (ROC AUC)}. 

\begin{definition}[ROC AUC] \label{def:ROCAUC}
    The \emph{receiver operating characteristic (ROC)} is a plot of false positive rates vs true positive rates of a binary classification method at various threshold values. 

    The \emph{receiver operating characteristic area under the curve (ROC AUC)} is the area under an ROC curve.
\end{definition}

Each model was trained for 20,000 epochs. Every 20 epochs, the ROC AUC achieved by the model on the NBIDE dataset was recorded. In Figure \ref{tab:max_results} we reported the maximum ROC AUC attained during training. We noticed that accuracy varied significantly between epochs, accordingly then smoothed our results by averaging the recorded ROC AUC scores over 200 epochs. For each model, the maximum smoothed ROC AUC attained during training is reported in Figure \ref{tab:avg_results}.

For each architecture, we trained five neural networks and averaged the maximum ROC AUC across all five networks. This result is given in the \textit{Avg High ROC AUC} column of \ref{tab:info}. Similarly, we averaged the smoothed maximum ROC AUCs across all five networks. The results are given in the \textit{Avg Smoothed High ROC AUC} column of \ref{tab:info}.

\subsection{Results}

\begin{figure}[htpb]
    \centering
    \begin{tabular}{|l|c|c|c|c|c|c|}
            \hline
            Method & Layers & Parameters & Avg High ROC AUC & Avg Smoothed High ROC AUC \\ \hline
            \textbf{Ref. Model} & 11 & 79,536 & \textbf{0.892} & \textbf{0.885} \\ \hline
            Double Block & 20 & 152,480 & 0.791 & 0.768\\ \hline
            Block Depth = 2 & 8 & 60,336 & 0.874 & 0.867 \\ \hline
            Block Depth = 4 & 14 & 128,032 & 0.813 & 0.800 \\ \hline
            Width = 8 & 11 & 20,248 & 0.841 & 0.830 \\ \hline
            Width = 32 & 11 & 315,232 & 0.878 & 0.869 \\ \hline
            Width = 64 & 11 & 1,205,244 & 0.875 & 0.865 \\ \hline
            CMC & NA & NA & 0.867 & 0.867 \\ \hline
        \end{tabular}
        \caption{Each model configuration was trained for 20,000 epochs on the E3 dataset. Every 20 epochs, the ROC AUC was recorded. A sliding average was taken over every 200 epochs, and the maximum attained average is reported here. For each model configuration, this process was repeated for a total of 5 runs.}
        \label{tab:info}
\end{figure}

Table \ref{tab:info} shows, for each model architecture used, the number of layers in the model and the number of parameters in the model. In the third column, the highest ROC AUC achieved during training by each architecture was averaged across all five training runs. The final column displays the average of the smoothed highest ROC AUC. Considering the overall Avg. High ROC AUC scores, 4 of our 7 model architectures outperformed the CMC algorithm. One of these architectures was our Reference Model, another was the same model with all of the width parameters doubled, and another was the same model with the width parameters doubled yet again. The final model which outperformed CMC was our `Block Depth 2' model (Figure \ref{fig:other_models} (C)). This model is similar to the Reference Model but with a 1x1 convolutional layer removed from each residual block. 

In deep learning, bigger is better \citep{amodei2016deep, sun2017revisiting}. Specifically, we mean that a larger model trained on a larger dataset performs better than a smaller model trained with a smaller dataset. With a training dataset of 2967 samples, we are not surprised that a model with 79,536 parameters outperforms one with 1,205,244. We expect that with more training data, all models will improve. We would also expect that with more training data, larger models would improve.  

In deep learning, the ratio between the number of parameters $p$ and size of the training dataset $n$ is often of interest. Our best-performing model architecture --the Reference model-- consists of $p_{ref}=79,536$ parameters and was trained on a dataset of $n_{ref} = 2,967$ images. The ratio of parameters to size of training set for our optimal model is $r_{ref}=\frac{p_{ref}}{n_{ref}}=26.81$. Comparatively, the Vision Transformer (ViT), \citep{dosovitskiy2020image}, the state-of-the-art architecture for the CIFAR-10 image classification task \citep{krizhevsky2009learning}, has $p_{ViT}=632,000,000$ parameters and was trained on a dataset of $n_{ViT} = 300,000,000$ images for a ratio of $r_{ViT}=\frac{p_{ViT}}{n_{ViT}}=2.11$. Reducing the number of parameters in our model resulted in worse generalization capability, suggesting that a smaller model may not have a large enough capacity to fulfill our task. Adding more parameters generally increases the risk of overfitting, but adding more training data generally decreases the risk of overfitting. Therefore, increasing the size of the training dataset would likely lead to a performance increase, as well as allow for training networks with more parameters.

\begin{figure}[htpb]
      \centering
      \includegraphics[width=.8\textwidth]{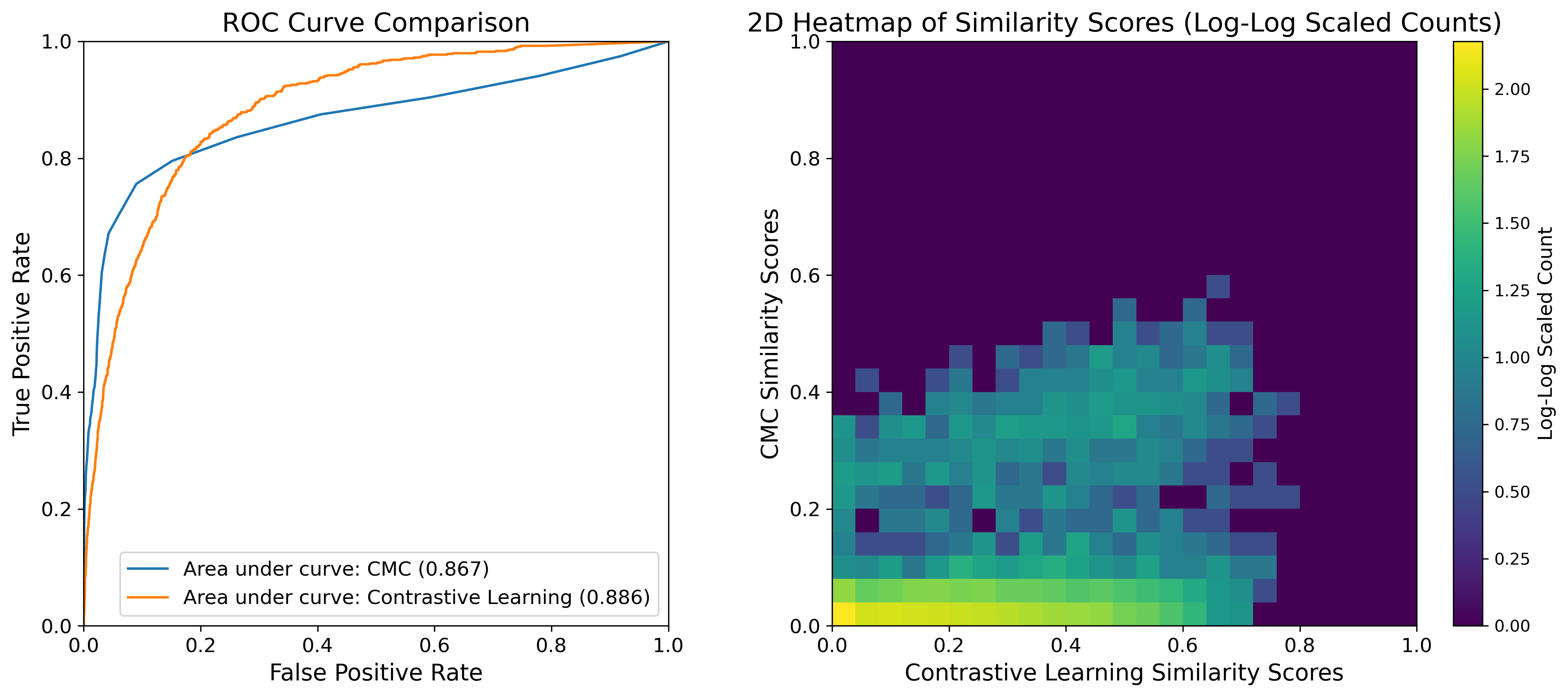}
      \caption{Left: ROC Curves generated by a contrastive neural network and CMC. The contrastive network used the `Width = 32' architecture and was trained for 20,000 epochs. Right: Heatmap of similarity scores obtained from contrastive learning (x-axis) and CMC (y-axis).}
      \label{fig:roc_heatmap}
\end{figure}

The ROC curves obtained by contrastive neural networks and CMC are shown on the left side of Figure \ref{fig:roc_heatmap}. Interestingly, contrastive learning yielded a higher AUC by being stronger in a different respect than CMC. On the one hand, while maintaining a low false positive rate, CMC can achieve a reasonably high true positive rate. On the other, contrastive learning achieves a near 100\% true positive rate with a considerably lower false positive rate than CMC. In terms of precision, contrastive learning begins to have higher precision than CMC at low match-classification threshold values. At high threshold values, CMC has high precision.

The plot on the right side of Figure \ref{fig:roc_heatmap} shows a heatmap of the scores produced by contrastive learning and CMC. We provided this heatmap to investigate whether the scores obtained by the two methods would be correlated. For non-zero CMC similarity scores, the two methods seem to produce correlated scores.

\begin{figure}[htpb]
      \centering
      \includegraphics[width=.8\textwidth]{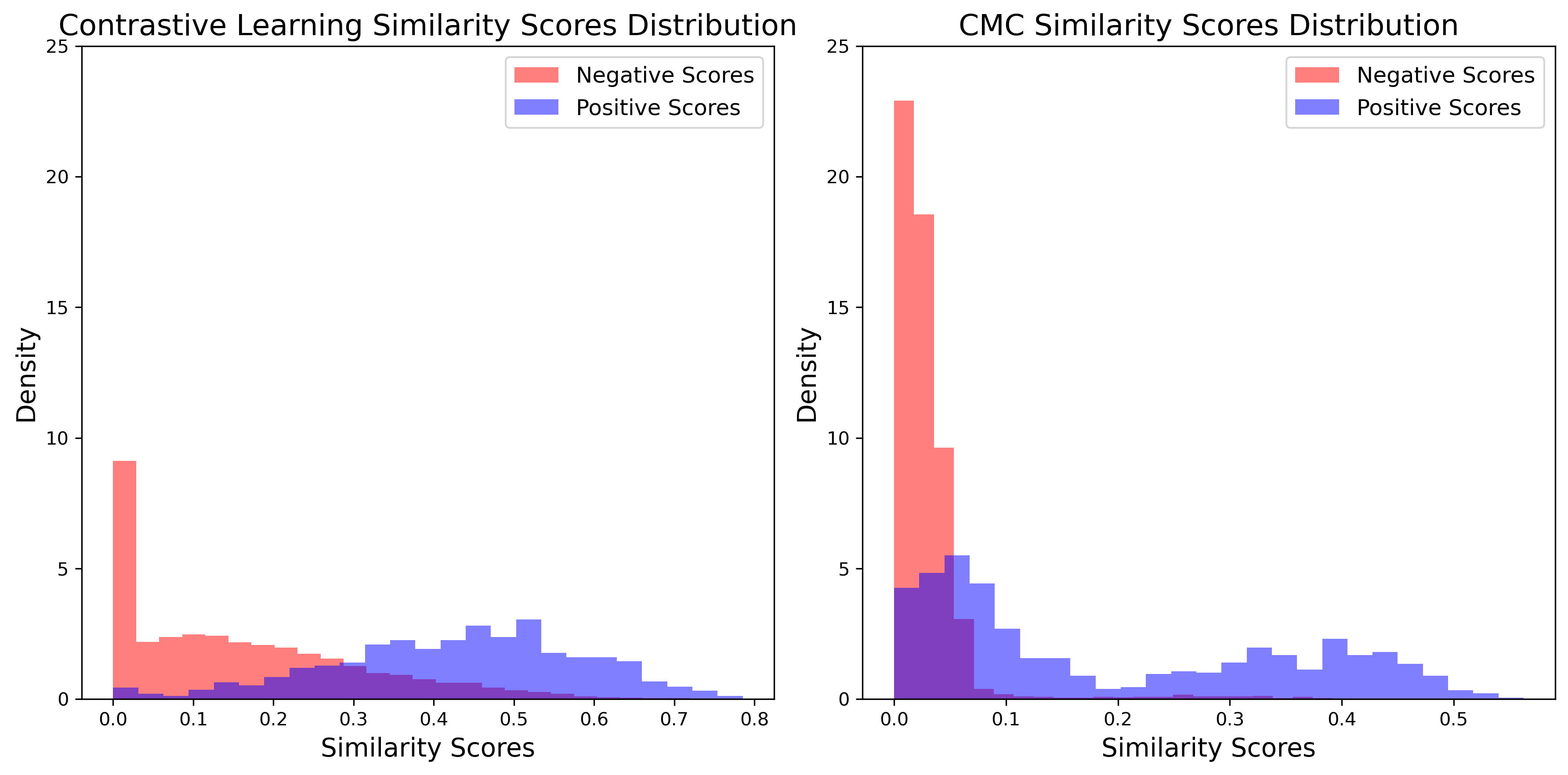}
      \caption{Histograms of similarity scores for cartridges in the NBIDE dataset. In red are the scores of cartridges fired from different guns. In blue are scores of cartridges fired from the same gun. The left histogram shows the scores from contrastive learning. The right histogram shows the scores from CMC.}
      \label{fig:histograms}
\end{figure}

The histograms of scores obtained by the two methods are shown in Figure \ref{fig:histograms}. On the left is the histogram of contrastive learning scores, and on the right are the CMC scores. In both histograms, the similarity scores between cartridges that were fired from the same gun are colored blue, and the scores between cartridges fired from different guns are colored red.

For contrastive learning, the negative scores are not nearly as concentrated around zero as the CMC. However, there is a high concentration at 0 and seemingly a second mode slightly above 0.1. These two modes are likely the negative scores between cartridges fired by different guns of different models and negative scores between different guns of the same model. Aside from the spike in negative scores at 0, there is a very smooth distribution of both positive and negative scores in contrastive learning. Because there is not a large number of positives distributed around 0 in contrastive learning, the method is able to achieve near 100\% true positive rate while maintaining a lower false positive rate than CMC.

Notice the negative scores for CMC are extremely concentrated around 0, which corresponds to the ability of CMC to maintain a low false positive rate at a reasonably high true positive rate. Interestingly, there seem to be two distributions of positive scores produced by CMC, one very low and one much higher. The low cluster may correspond to scores between cartridges from the same manufacturer while high cartridges from different manufacturers.

\section{Discussion}

Contrastive learning achieved a higher ROC AUC than the CMC algorithm; however, the ROC curves in Figure \ref{fig:roc_heatmap} show that the two methods have their own strengths. Contrastive learning can achieve an almost perfect true positive rate with a much lower false positive rate than CMC. Meanwhile, CMC can attain a considerable true positive rate while maintaining a negligible false positive rate. Therefore the results of our experiments show that contrastive learning is probably more useful at low threshold values, when the value of a true positive is greater than the cost of a false positive. Conversely, CMC is probably more useful when the cost of a false positive is greater than the value of a true positive. However, with larger contrastive neural networks and trained on larger datasets, the limitations of contrastive learning may be overcome. 

Comparing the method of contrastive learning with the CMC algorithm is complicated by the need to train the contrastive neural networks and not the CMC algorithm. We remark that, although CMC does not have parameters that need to be trained, it does have hyperparameters that needs to be carefully calibrated (translation, rotation, and correlation thresholds discussed above). The quality of the calibration has a considerable effect on overall performance. One advantage of a deep learning approach is avoiding the need for this calibration.

The contrastive learning approach was significantly faster than the CMC algorithm, even though training a neural network is an overhead cost not incurred by CMC. For contrastive learning our largest contrastive networks trained for less than 12 hours on a dataset of 2967 cartridge casing scans, and their peak performance was usually attained within the first 2 hours. After training, testing the contrastive network on the NBIDE dataset of 144 scans is nearly instant. For CMC --outside of tweaking the hyperparameters from their default values in \citet{zemmels2023cartridge}-- the CMC algorithm is not trained. However, parallelized across 48 CPUs, it took over 9 days for the CMC algorithm to determine similarity scores between all 144 cartridge casings in the NBIDE dataset. 

\subsection{False Starts}

Our initial attempts to use the NBIDE dataset for both training and testing with leave-two-out cross-validation. In this setup, the contrastive network was trained on 120 cartridges fired by 10 of the firearms and tested on the 24 cartridges from the remaining 2 firearms. The approach was not competitive with the CMC algorithm, presumably because of the small size training dataset. Only after including the E3 dataset of 2967 cartridges in training, was contrastive learning able to outperform CMC. 

The performance gains from a larger training dataset encouraged us to incorporate a second training dataset. This second set was the Popstat dataset, which is an aggregate of cartridge casing scans from other datasets. Because of this, we had to remove all cartridges from the Popstat dataset that are contained in our testing NBIDE dataset. This left us with 350 new images of cartridge casings, fired by 175 guns, each gun contributing two cartridge casings. Surprisingly, when training a contrastive network on the aggregated E3 and Popstat datasets, we achieved worse performance than simply training on the E3 dataset. Our intuition is that because the Popstat dataset only contains 2 exemplars from each firearm, the optimization process heavily incentivizes the network to declare cartridge casings to be dissimilar. We conclude from this, that a new and effective dataset would contain many cartridges fired from each firearm present. 

\subsection{Future Work}

 In this work, we trained neural networks until they had overfit and showed that throughout the training process the networks achieved a greater ROC AUC than the CMC algorithm. To determine where in the training process to stop, the traditional machine learning motif of "bias-variance tradeoff" must be tended too. 

 We suspect that contrastive networks would achieve a greater ROC AUC if they were provided with both the breech face and the firing pin impression as input. Assuming so, the step of cropping out the firing pin would be an unnecessary source of bias in an otherwise automatable process. Therefore work investigating the ability of contrastive networks to simultaneously discriminate firing pin impressions and breech faces could allow for the removal of superfluous preprocessing steps. Preprocessing will probably always be needed to some extent, but the more it can be avoided or at least automated in an unbiased way, the better. 

Our results demonstrate that, with a training dataset of only 2967 samples, contrastive networks achieve competitive results to the widely adopted CMC algorithm. A generally held principle in the field of deep learning is that a larger dataset yields a better model \citep{amodei2016deep, sun2017revisiting}. Unfortunately, the E3 dataset was the largest readily available dataset on the NIST Ballistics Toolmarks Database. The introduction of a larger dataset of high-quality 3D cartridge casing scans would likely allow for the training of neural networks to a much higher degree of accuracy. 

\section{Conclusion}

Our results suggest that contrastive neural networks are capable of outperforming existing statistical methods for the task of comparing cartridge casing scans. Specifically, we compared contrastive neural networks to the Congruent Matching Cells (CMC) algorithm. With a training dataset of 2967 images, we were able to train contrastive networks that achieve a higher ROC AUC than the CMC algorithm on the NBIDE dataset. Our findings support the use of contrastive neural networks to assist practitioners in the examination of forensic evidence such as cartridge casings. 

The results of our ablation study indicated moderate robustness of our neural networks to changes in architecture. The introduction of larger datasets of cartridge casing scans would likely allow for the training of larger, more effective networks. 

\section{Acknowledgments}

Cole Patten received support from CAPITAL Services Inc. and the National Science Foundation through the National Research Traineeship site titled “Cyber-Physical-Social System for Understanding and Thwarting the Illicit Economy” under NSF award numbers: DGE 1828492, 1828302, 1828462, 1828288. Dr. Christopher Saunders received support from Award No. 15PNIJ-23-GG-04232-MUMU, awarded by the National Institute of Justice, Office of Justice Programs, U.S. Department of Justice. Michael Puthawala received support from CAPITAL Services Inc.

The opinions, findings, and conclusions or recommendations expressed in this publication are those of the authors and do not necessarily reflect the official position or policies of the Department of Justice.

\bibliography{references}

\newpage
\begin{appendices}
\section{First Appendix}

\begin{figure}[htpb]
      \centering
      \includegraphics[width=.75\textwidth]{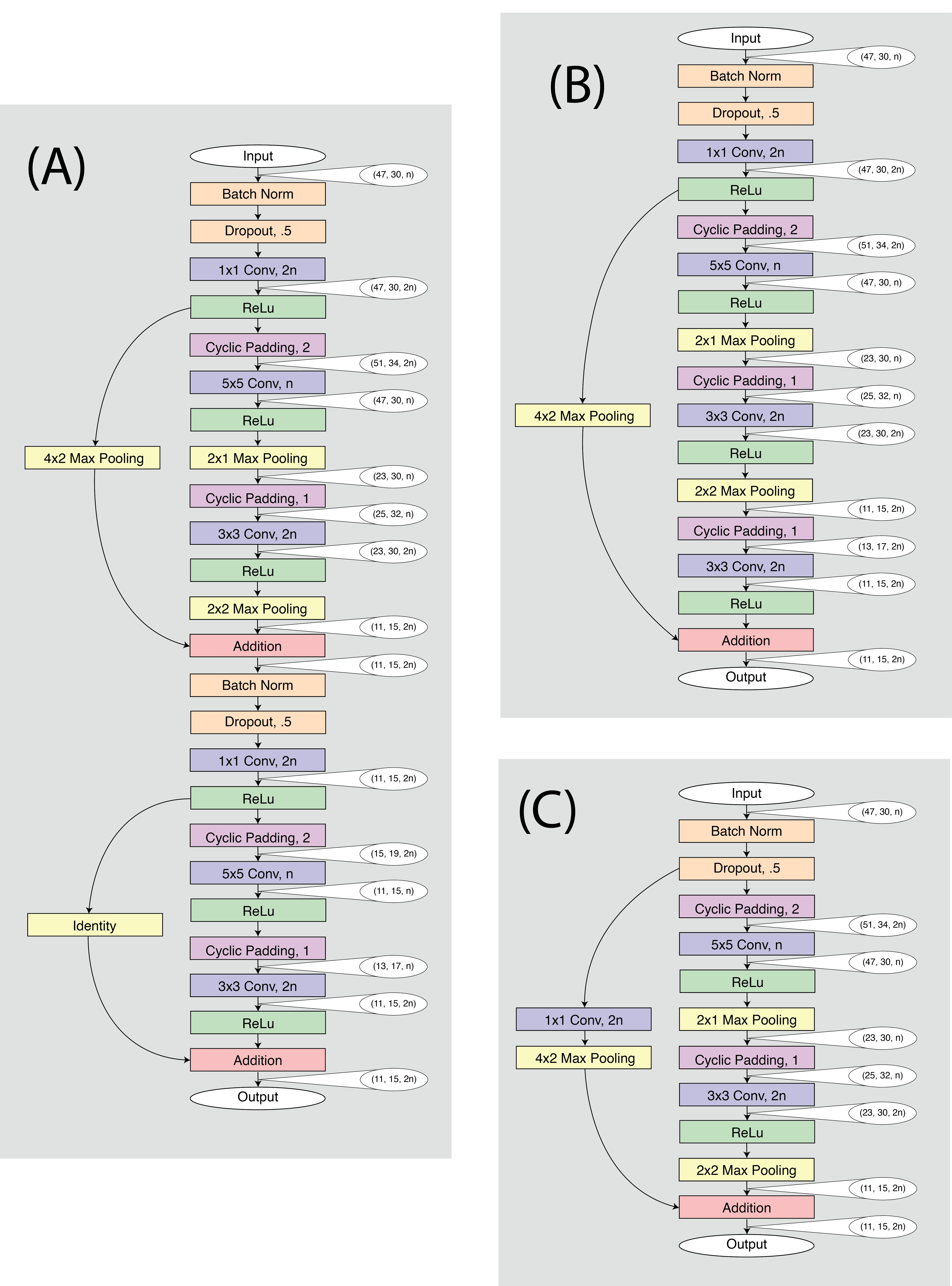}
      \caption{(A) The residual block of our ``Double Block'' model. Compared to the ``Reference Model'', each residual block is repeated with the pooling layers omitted. (B) The residual block of our ``Block Depth = 4'' model. Compared to the ``Reference Model'', each residual block has an additional cyclic padding, convolutional layer, and ReLu activation. (C) The residual block of our ``Block Depth = 2'' model. Compared to the ``Reference Model'', the 1x1 convolution at the beginning of each block is moved within the ``skip connection'' and is not followed by a non-linear activation. }
      \label{fig:other_models}
\end{figure}

\begin{figure}[htpb]
    \centering
        \begin{minipage}[t]{0.49\textwidth}
        \centering
        \resizebox{\textwidth}{!}{%
            \begin{tabular}{|l|c|c|c|c|c|c|}
                \hline
                Method & Run 1 & Run 2 & Run 3 & Run 4 & Run 5 & Mean \\ \hline
                \textbf{Ref. Model} & 0.897 & 0.873 & 0.899 & 0.901 & 0.890 & \textbf{0.892} \\ \hline
                Double Block & 0.780 & 0.778 & 0.836 & 0.796 & 0.764 & 0.791 \\ \hline
                Block Depth = 2 & 0.863 & 0.888 & 0.885 & 0.860 & 0.875 & 0.874 \\ \hline
                Block Depth = 4 & 0.830 & 0.801 & 0.831 & 0.808 & 0.797 & 0.813 \\ \hline
                Width = 8 & 0.898 & 0.821 & 0.834 & 0.811 & 0.840 & 0.841 \\ \hline
                Width = 32 & 0.874 & 0.871 & 0.880 & 0.880 & 0.887 & 0.878 \\ \hline
                Width = 64 & 0.862 & 0.894 & 0.880 & 0.877 & 0.863 & 0.875 \\ \hline
                CMC & NA & NA & NA & NA & 0.867 & 0.867 \\ \hline
            \end{tabular}
        }
        \caption{Each model configuration was trained for 20,000 epochs on the E3 dataset. Every 20 epochs, the ROC AUC was recorded. The maximum attained ROC AUC is reported here. For each model configuration, this process was repeated for a total of 5 runs.}
        \label{tab:max_results}
    \end{minipage}
    \hfill
    \begin{minipage}[t]{0.49\textwidth}
        \centering
        \resizebox{\textwidth}{!}{%
            \begin{tabular}{|l|c|c|c|c|c|c|}
                \hline
                Method & Run 1 & Run 2 & Run 3 & Run 4 & Run 5 & Mean \\ \hline
                \textbf{Ref. Model} & 0.889 & 0.867 & 0.891 & 0.894 & 0.883 & \textbf{0.885} \\ \hline
                Double Blocks & 0.751 & 0.762 & 0.813 & 0.763 & 0.752 & 0.768 \\ \hline
                Block Depth = 2 & 0.856 & 0.879 & 0.882 & 0.850 & 0.868 & 0.867 \\ \hline
                Block Depth = 4 & 0.820 & 0.792 & 0.823 & 0.794 & 0.792 & 0.800 \\ \hline
                Width = 8 & 0.891 & 0.812 & 0.828 & 0.789 & 0.832 & 0.830 \\ \hline
                Width = 32 & 0.864 & 0.864 & 0.873 & 0.871 & 0.875 & 0.869 \\ \hline
                Width = 64 & 0.850 & 0.880 & 0.875 & 0.868 & 0.852 & 0.865 \\ \hline
                CMC & NA & NA & NA & NA & 0.867 & 0.867 \\ \hline
            \end{tabular}
        }
        \caption{Each model configuration was trained for 20,000 epochs on the E3 dataset. Every 20 epochs, the ROC AUC was recorded. A sliding average was taken over every 200 epochs, and the maximum attained average is reported here. For each model configuration, this process was repeated for a total of 5 runs.}
        \label{tab:avg_results}
    \end{minipage}
\end{figure}

\begin{figure}[htpb]
    \centering
    \hspace*{\fill} 
    
    \begin{minipage}[t]{0.45\textwidth} 
        \centering
        \includegraphics[width=\textwidth]{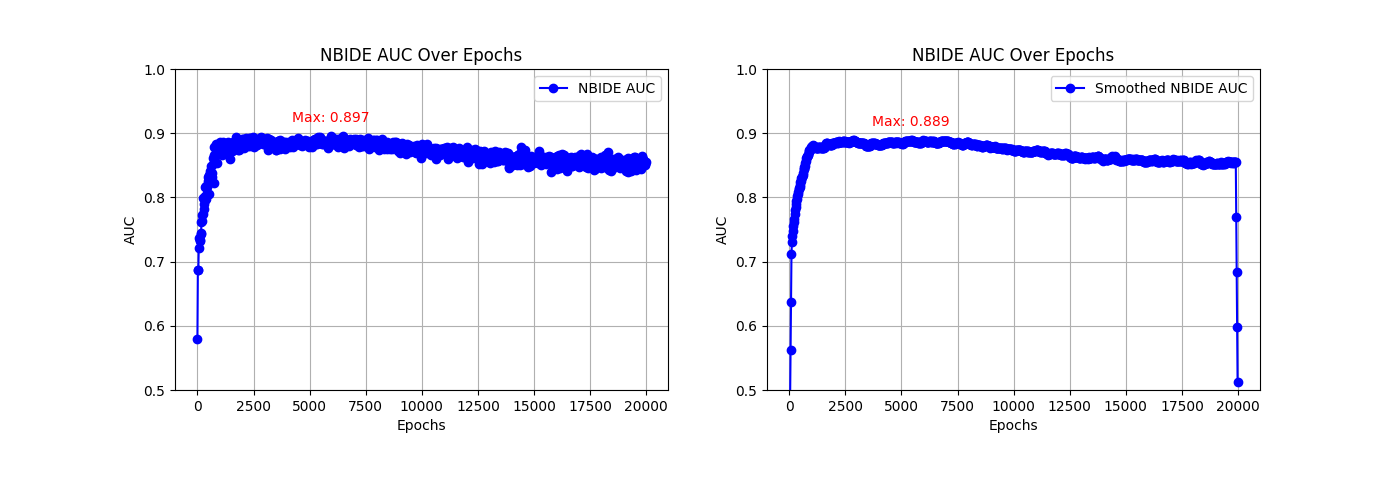}
        \caption*{(a) Reference Model first run.}
        \includegraphics[width=\textwidth]{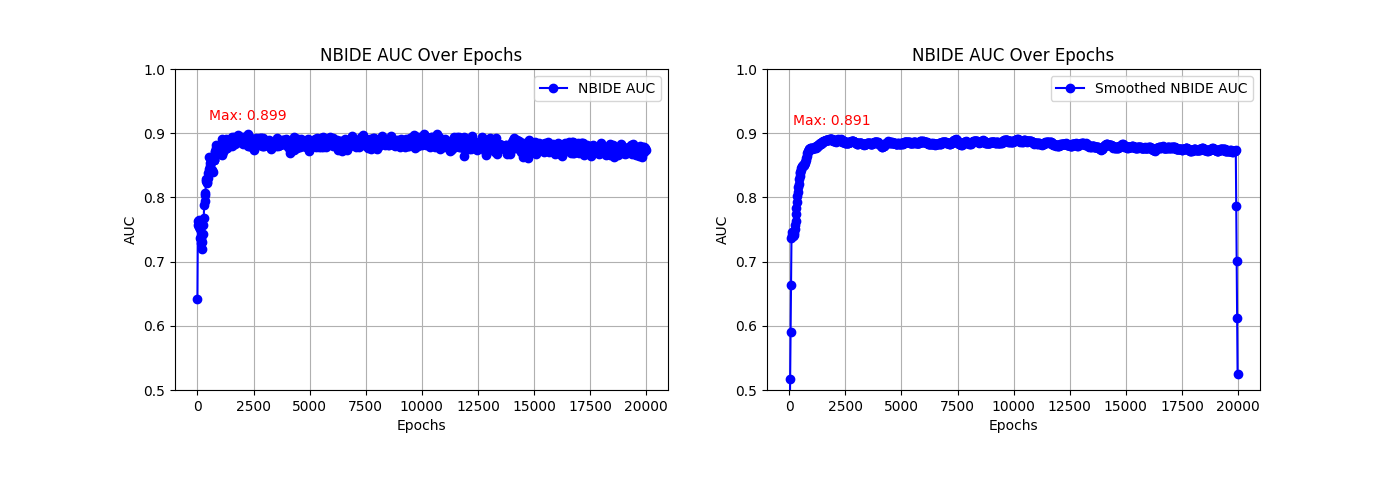}
        \caption*{(c) Reference Model third run.}
    \end{minipage}
    \hfill
    \begin{minipage}[t]{0.45\textwidth} 
        \centering
        \includegraphics[width=\textwidth]{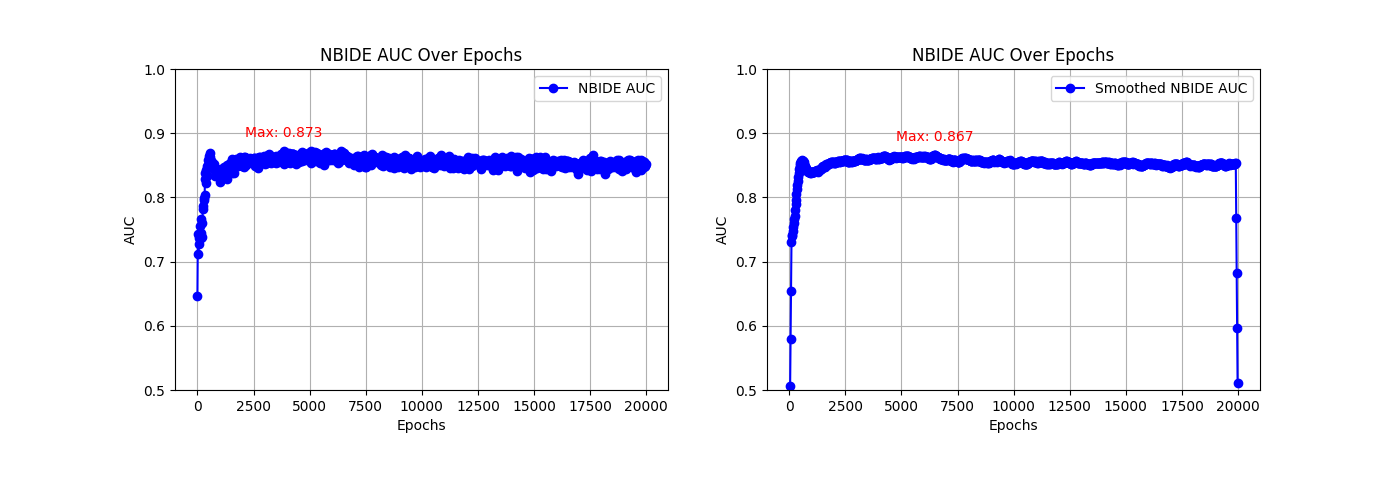}
        \caption*{(b) Reference Model second run.}
        \includegraphics[width=\textwidth]{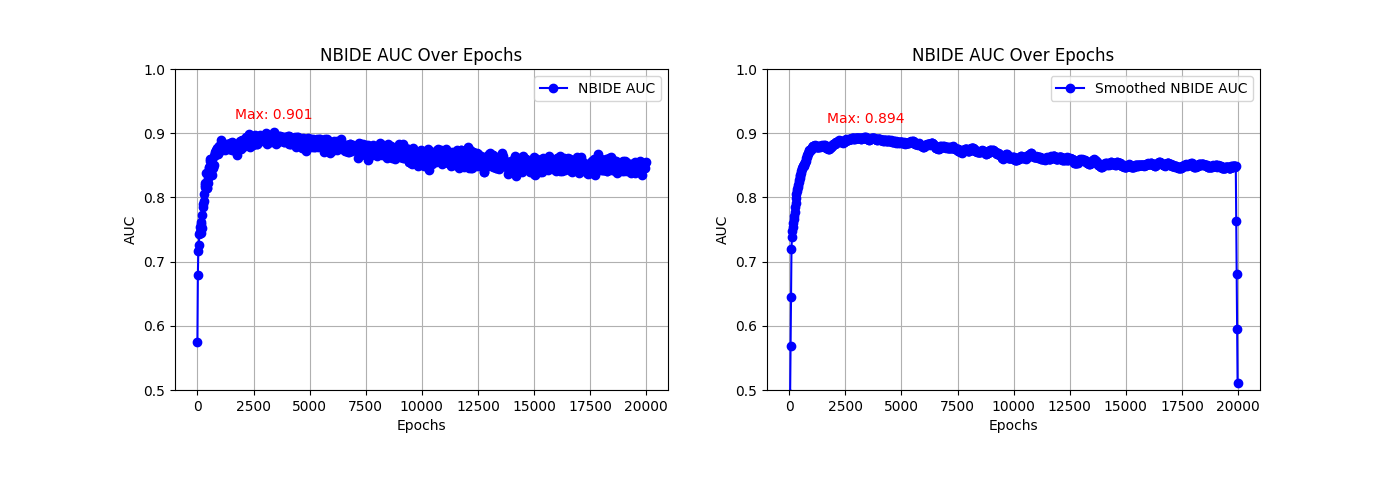}
        \caption*{(d) Reference Model fourth run.}
    \end{minipage}
    
    \hspace*{\fill} 

    \vspace{.8em} 
    \begin{minipage}[t]{0.45\textwidth}
        \centering
        \includegraphics[width=\textwidth]{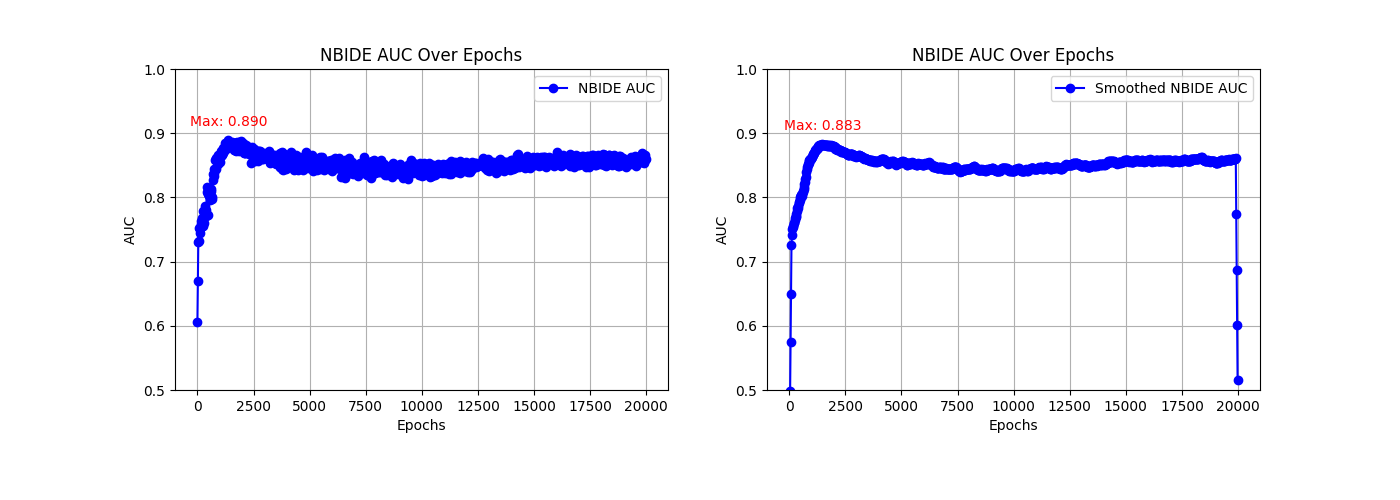}
        \caption*{(e) Reference Model fifth run.}
    \end{minipage}

    \caption{Plots of the ROC AUC attained by the Reference Model during 20,000 training epochs.}
    \label{fig:ref_plots}
\end{figure}

\begin{figure}[htpb]
    \centering
    \hspace*{\fill} 
    
    \begin{minipage}[t]{0.45\textwidth} 
        \centering
        \includegraphics[width=\textwidth]{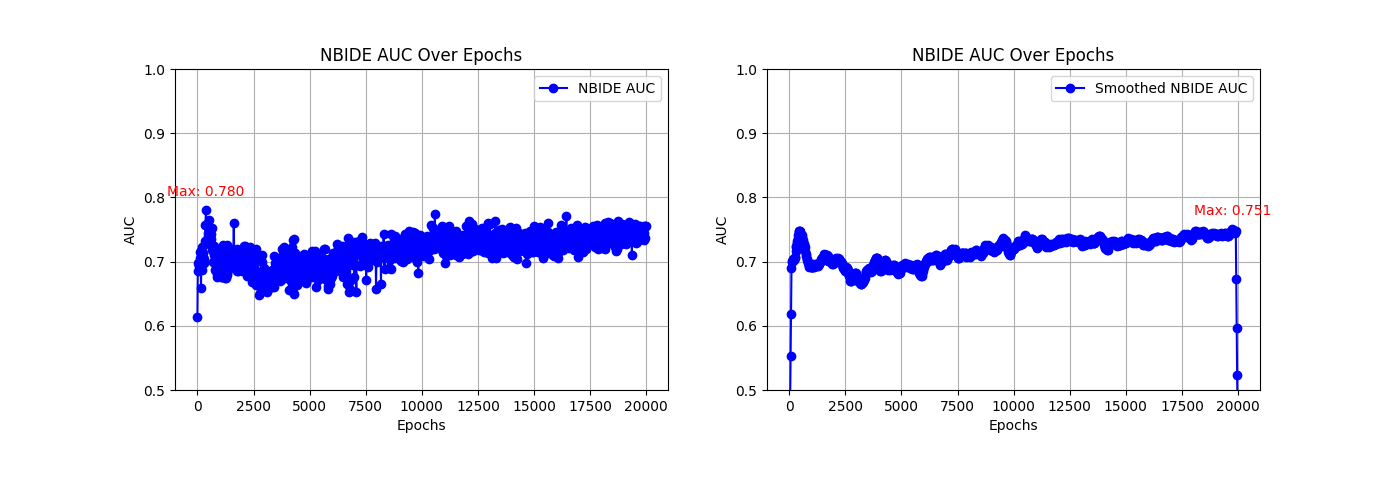}
        \caption*{(a) Double Block model first run.}
        \includegraphics[width=\textwidth]{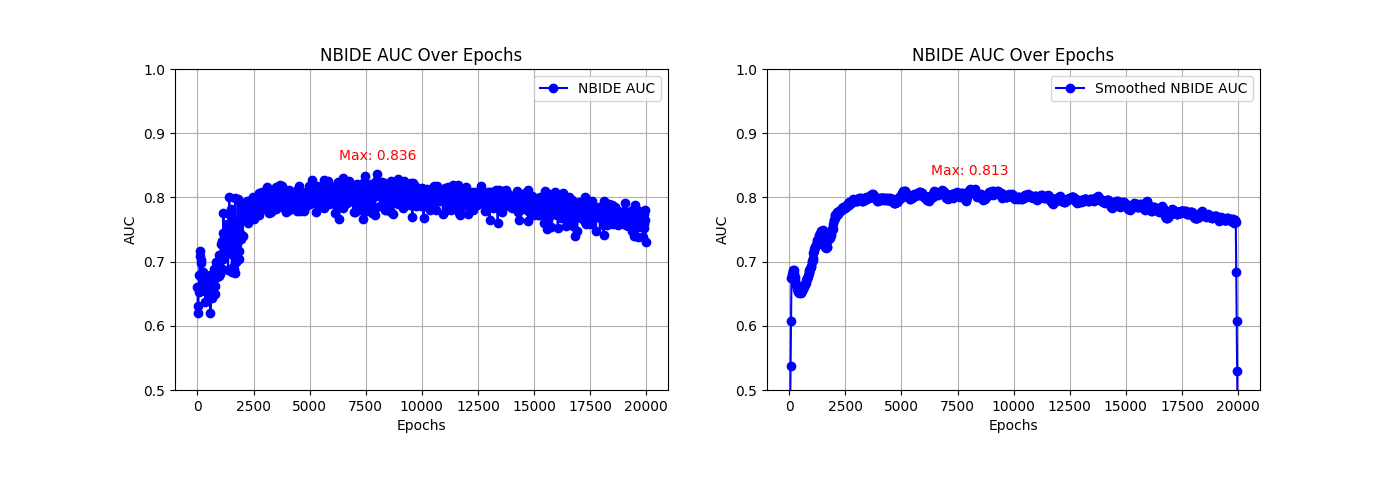}
        \caption*{(c) Double Block model third run.}
    \end{minipage}
    \hfill
    \begin{minipage}[t]{0.45\textwidth} 
        \centering
        \includegraphics[width=\textwidth]{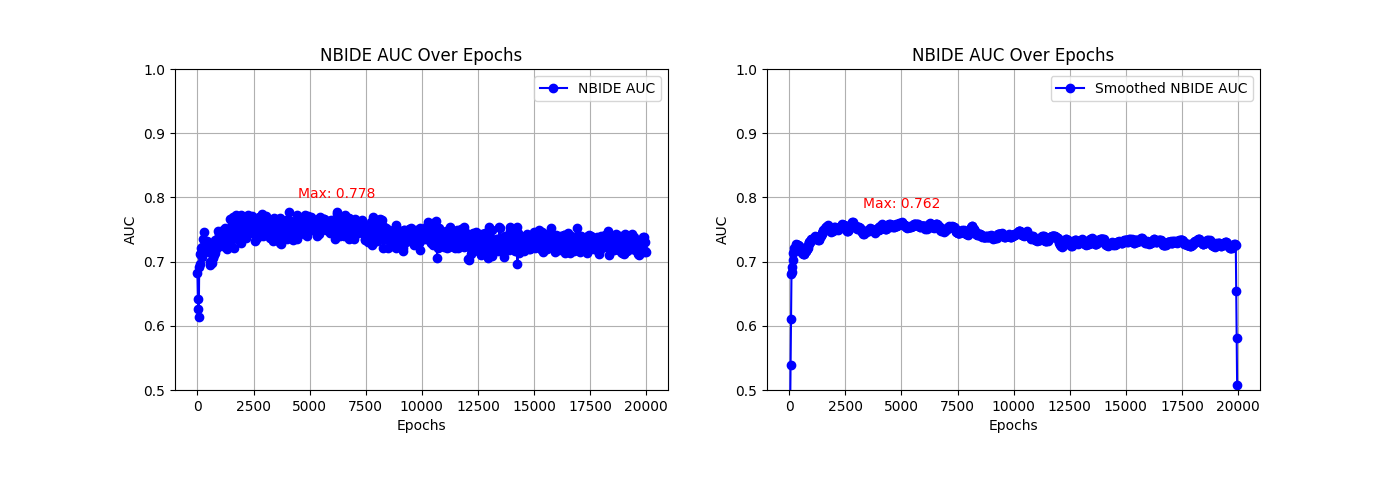}
        \caption*{(b) Double Block model second run.}
        \includegraphics[width=\textwidth]{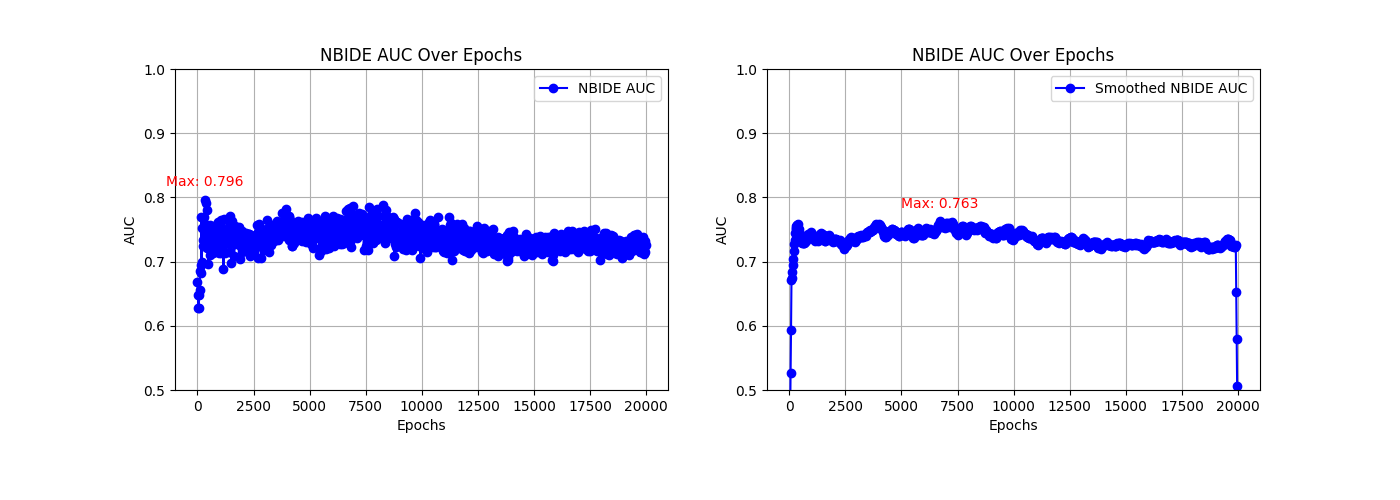}
        \caption*{(d) Double Block model fourth run.}
    \end{minipage}
    
    \hspace*{\fill} 

    \vspace{.8em} 
    \begin{minipage}[t]{0.45\textwidth}
        \centering
        \includegraphics[width=\textwidth]{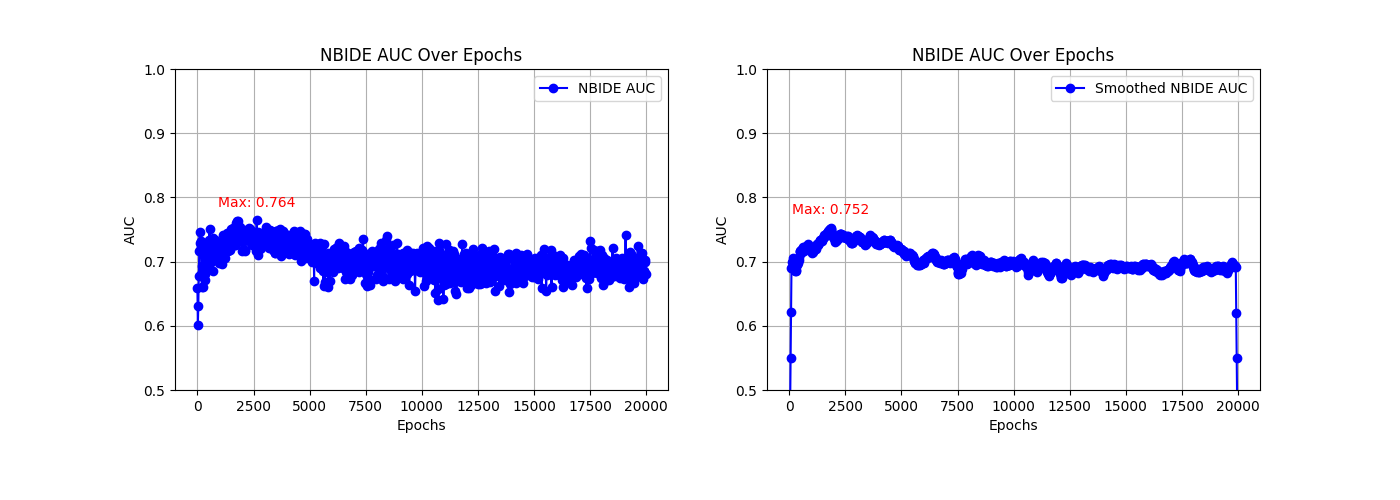}
        \caption*{(e) Double Block model fifth run.}
    \end{minipage}

    \caption{Plots of the ROC AUC attained by the Double Block model during 20,000 training epochs.}
    \label{fig:double_block_plots}
\end{figure}

\begin{figure}[htpb]
    \centering
    \hspace*{\fill} 
    
    \begin{minipage}[t]{0.45\textwidth} 
        \centering
        \includegraphics[width=\textwidth]{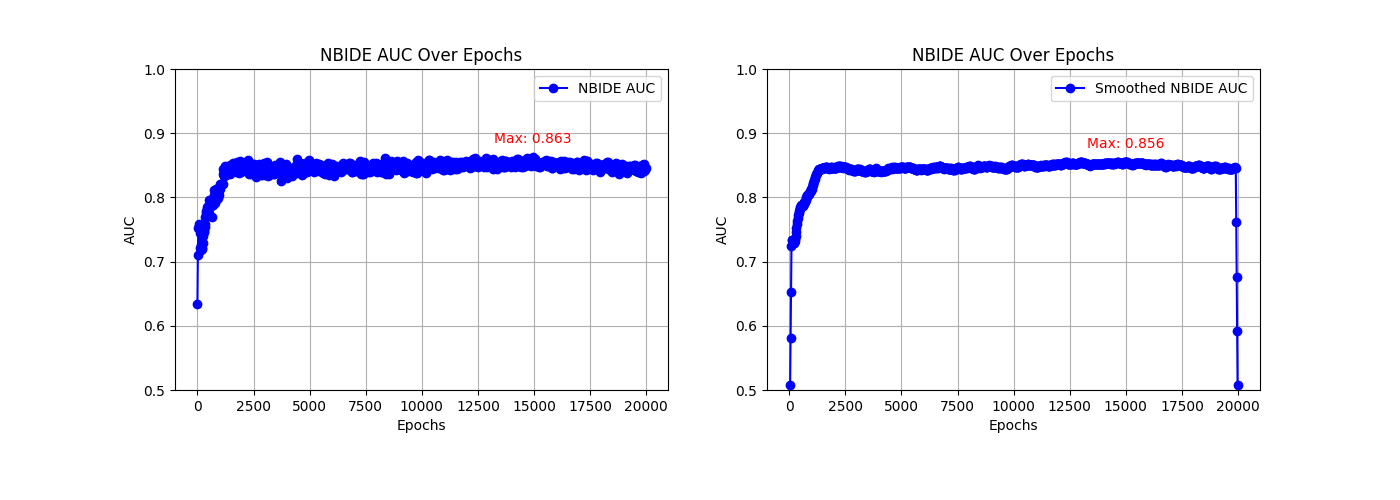}
        \caption*{(a) Block Depth=2 model first run.}
        \includegraphics[width=\textwidth]{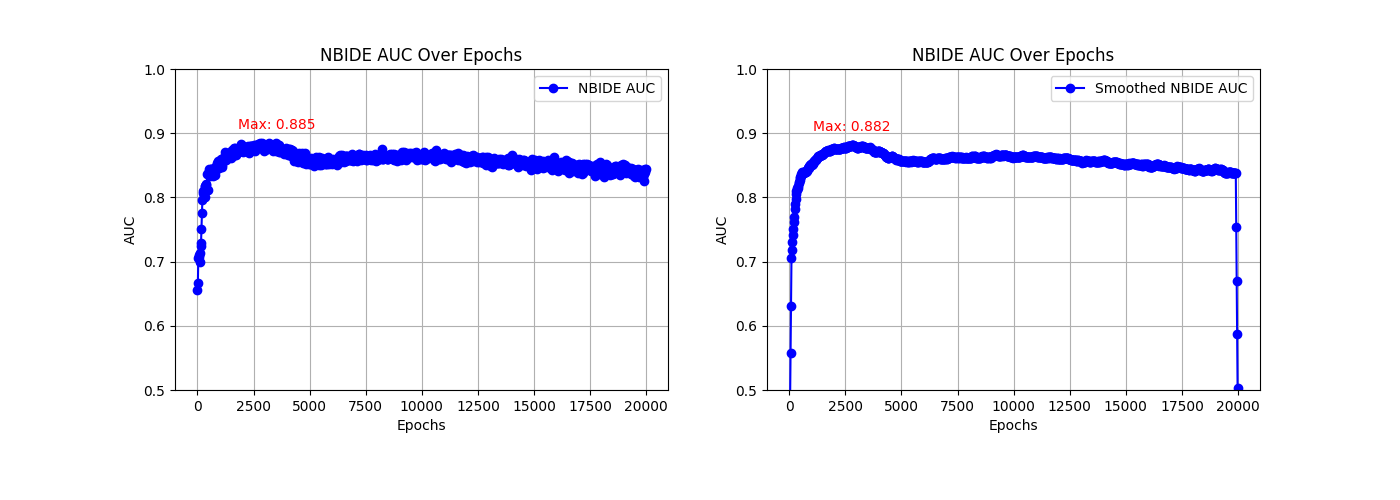}
        \caption*{(c) Block Depth=2 model third run.}
    \end{minipage}
    \hfill
    \begin{minipage}[t]{0.45\textwidth} 
        \centering
        \includegraphics[width=\textwidth]{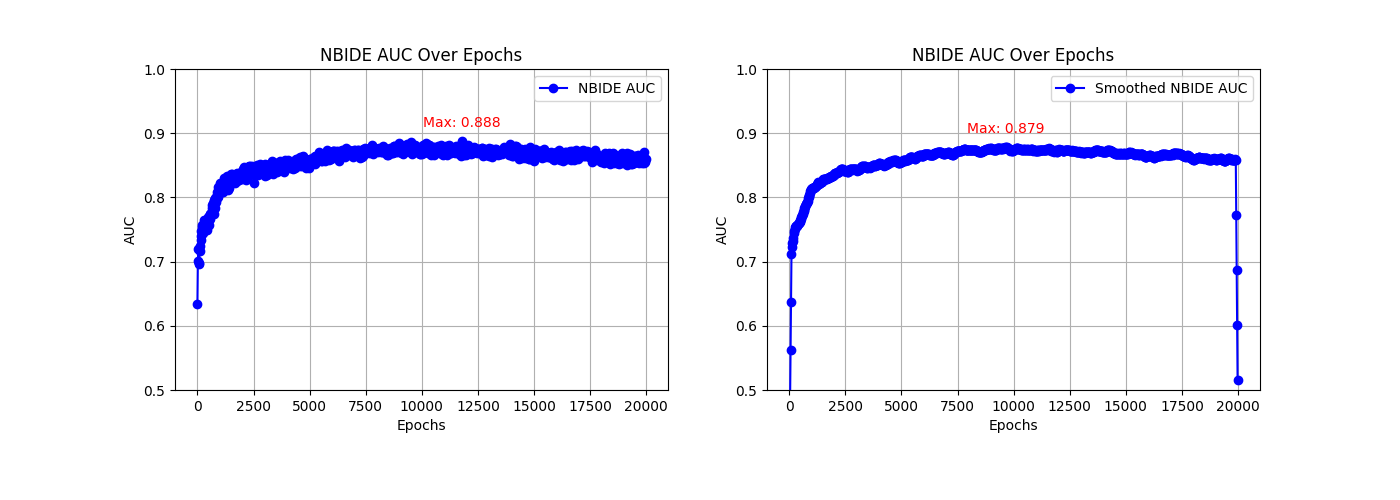}
        \caption*{(b) Block Depth=2 model second run.}
        \includegraphics[width=\textwidth]{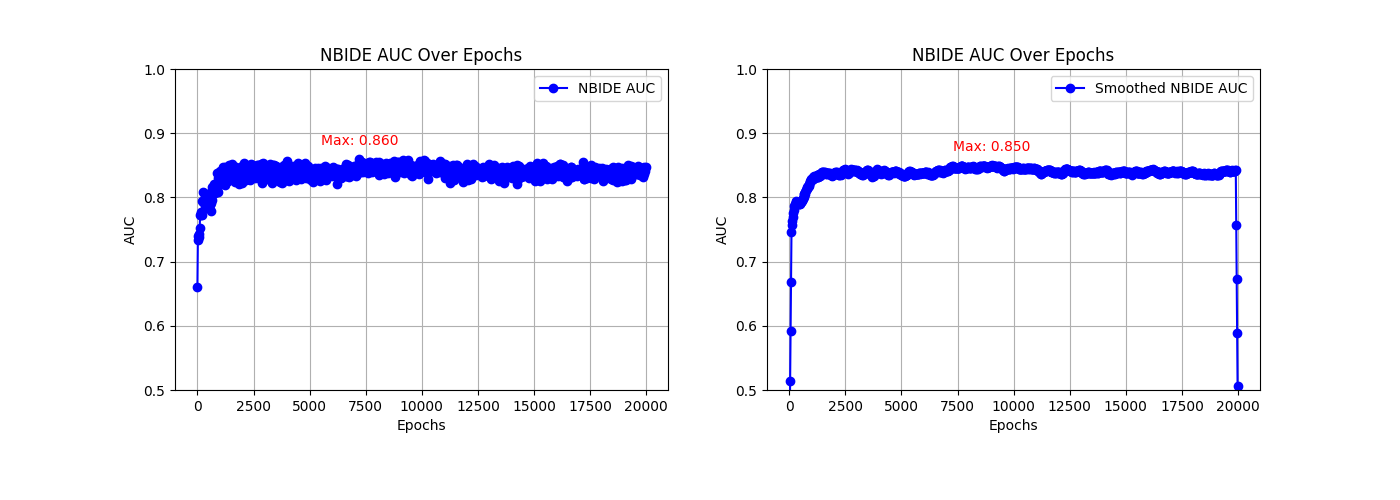}
        \caption*{(d) Block Depth=2 model fourth run.}
    \end{minipage}
    
    \hspace*{\fill} 

    \vspace{.8em} 
    \begin{minipage}[t]{0.45\textwidth}
        \centering
        \includegraphics[width=\textwidth]{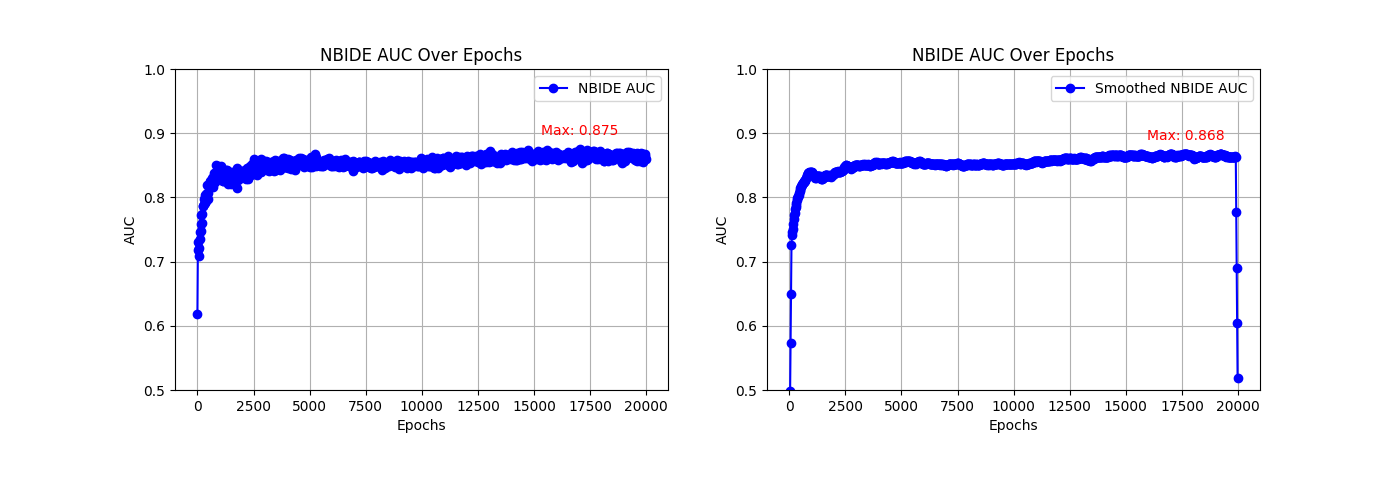}
        \caption*{(e) Block Depth=2 model fifth run.}
    \end{minipage}

    \caption{Plots of the ROC AUC attained by the Block Depth=2 model during 20,000 training epochs.}
    \label{fig:block_depth_2_plots}
\end{figure}

\begin{figure}[htpb]
    \centering
    \hspace*{\fill} 
    
    \begin{minipage}[t]{0.45\textwidth} 
        \centering
        \includegraphics[width=\textwidth]{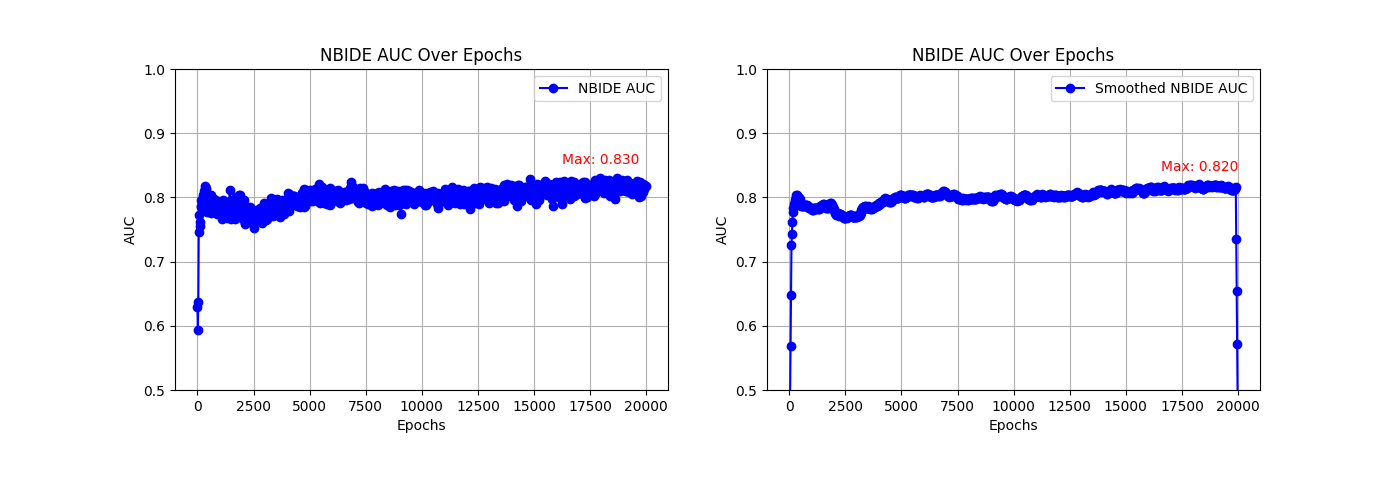}
        \caption*{(a) Block Depth=4 model first run.}
        \includegraphics[width=\textwidth]{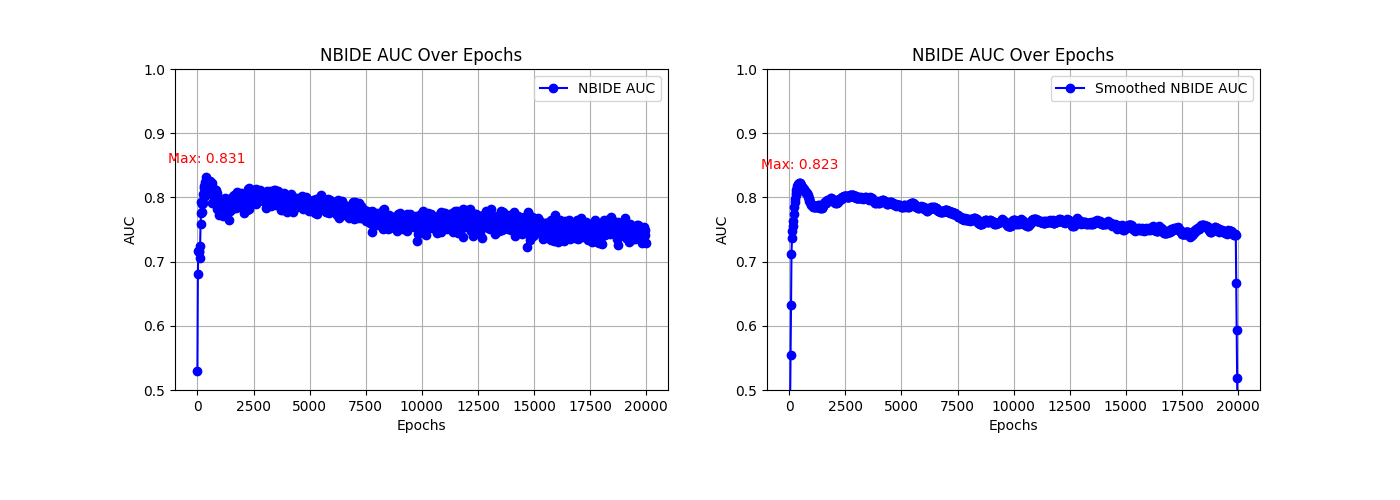}
        \caption*{(c) Block Depth=4 model third run.}
    \end{minipage}
    \hfill
    \begin{minipage}[t]{0.45\textwidth} 
        \centering
        \includegraphics[width=\textwidth]{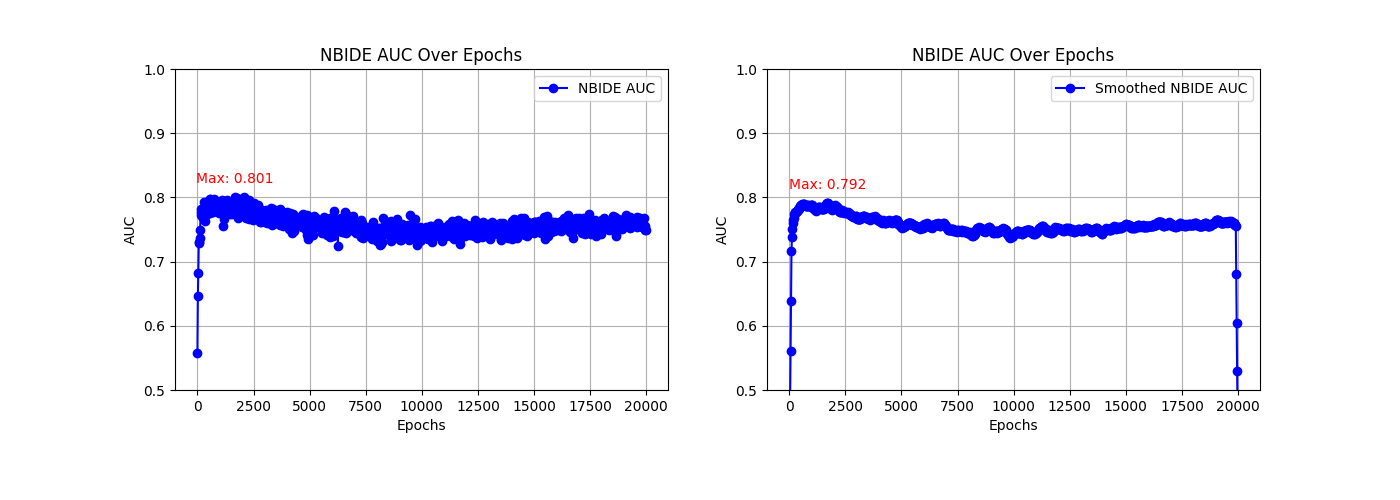}
        \caption*{(b) Block Depth=4 model second run.}
        \includegraphics[width=\textwidth]{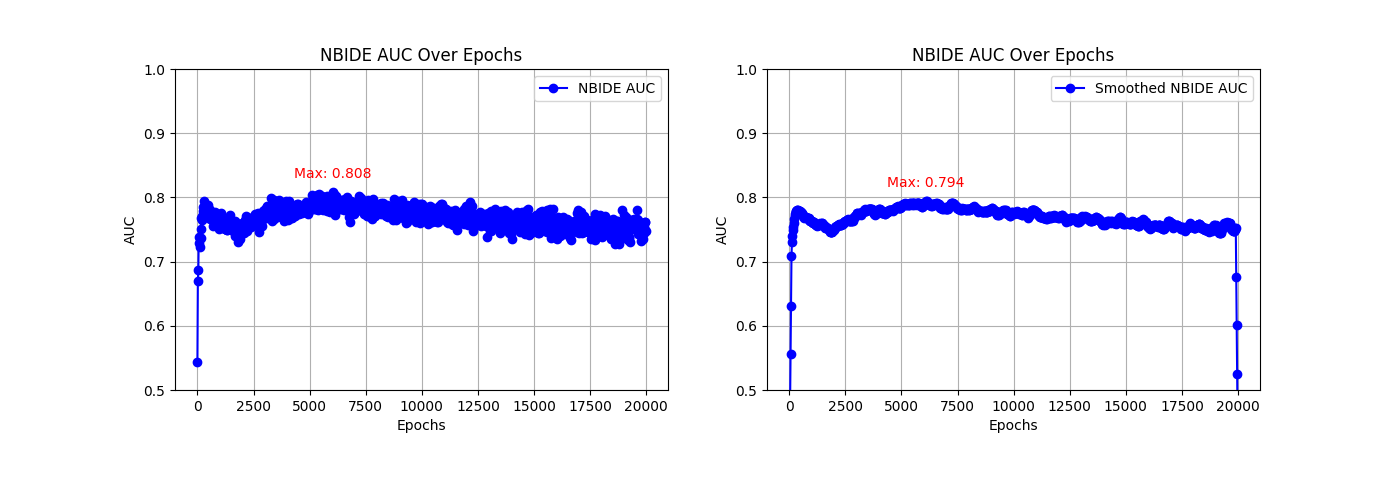}
        \caption*{(d) Block Depth=4 model fourth run.}
    \end{minipage}
    
    \hspace*{\fill} 

    \vspace{.8em} 
    \begin{minipage}[t]{0.45\textwidth}
        \centering
        \includegraphics[width=\textwidth]{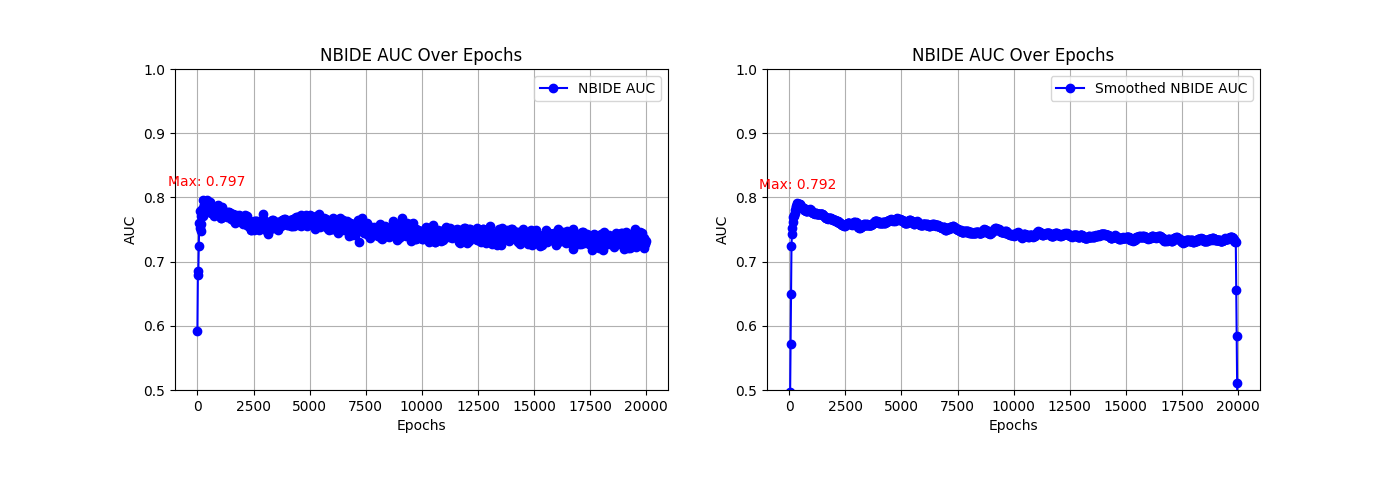}
        \caption*{(e) Block Depth=4 model fifth run.}
    \end{minipage}

    \caption{Plots of the ROC AUC attained by the Block Depth=4 model during 20,000 training epochs.}
    \label{fig:block_depth_4_plots}
\end{figure}

\begin{figure}[htpb]
    \centering
    \hspace*{\fill} 
    
    \begin{minipage}[t]{0.45\textwidth} 
        \centering
        \includegraphics[width=\textwidth]{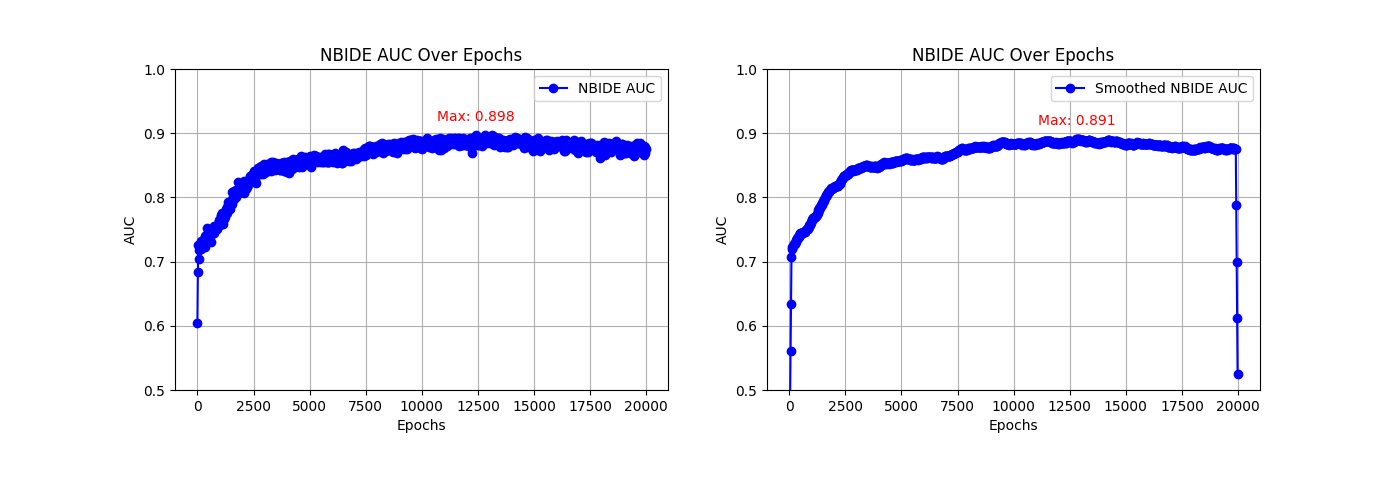}
        \caption*{(a) Width=8 model first run.}
        \includegraphics[width=\textwidth]{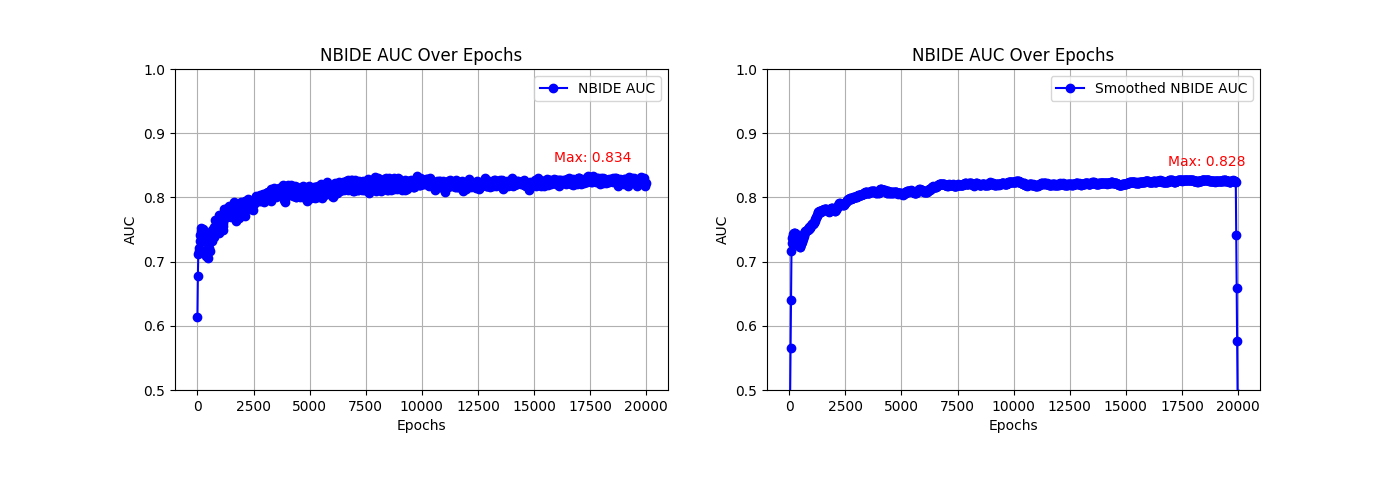}
        \caption*{(c) Width=8 model third run.}
    \end{minipage}
    \hfill
    \begin{minipage}[t]{0.45\textwidth} 
        \centering
        \includegraphics[width=\textwidth]{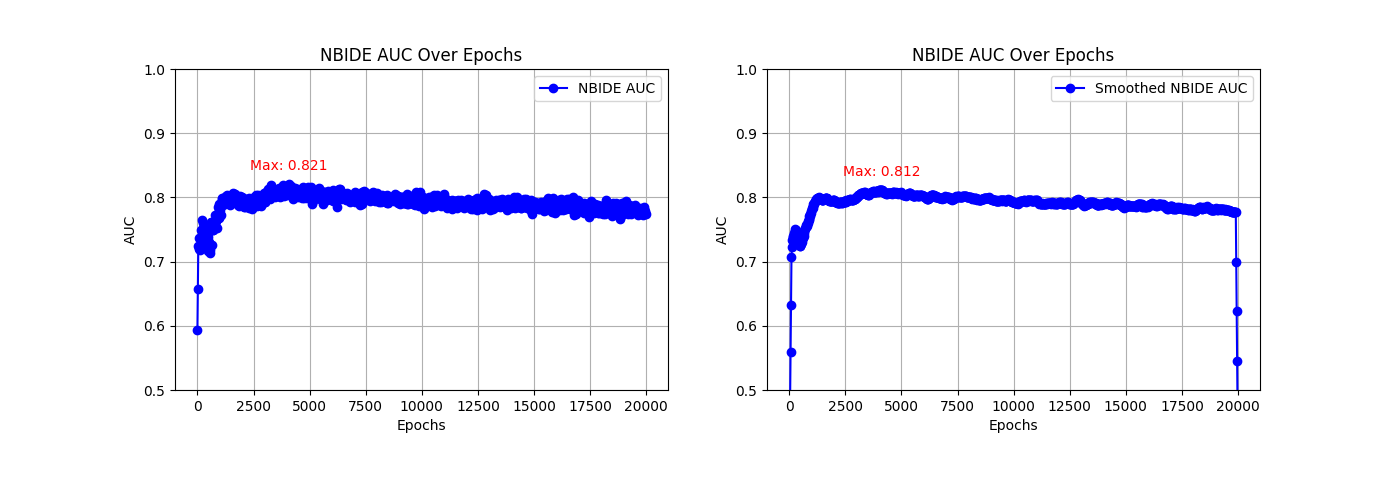}
        \caption*{(b) Width=8 model second run.}
        \includegraphics[width=\textwidth]{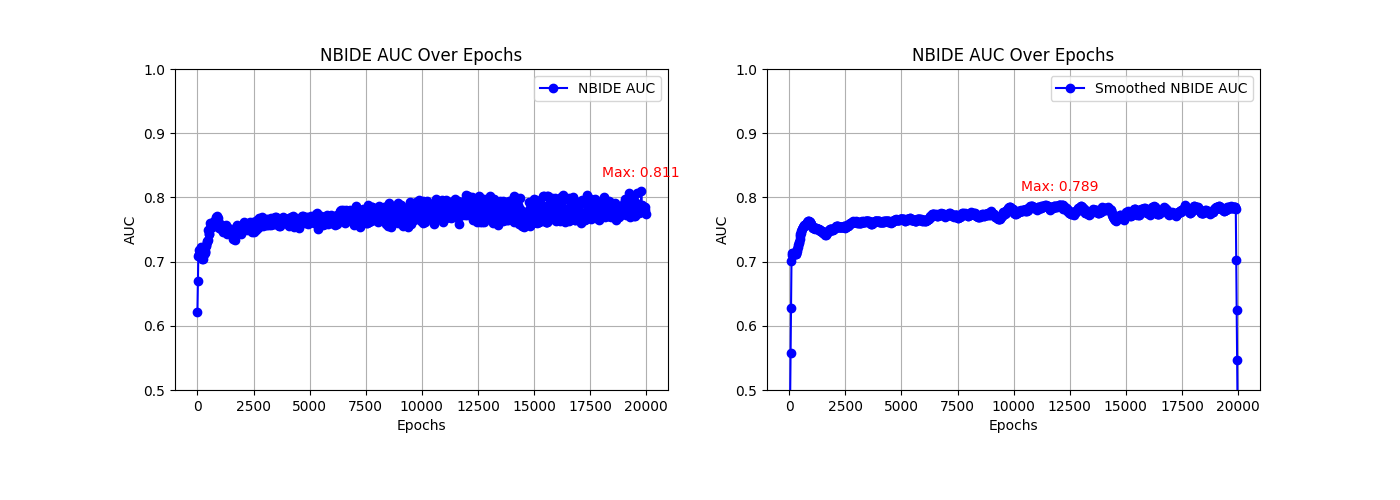}
        \caption*{(d) Width=8 model fourth run.}
    \end{minipage}
    
    \hspace*{\fill} 

    \vspace{.8em} 
    \begin{minipage}[t]{0.45\textwidth}
        \centering
        \includegraphics[width=\textwidth]{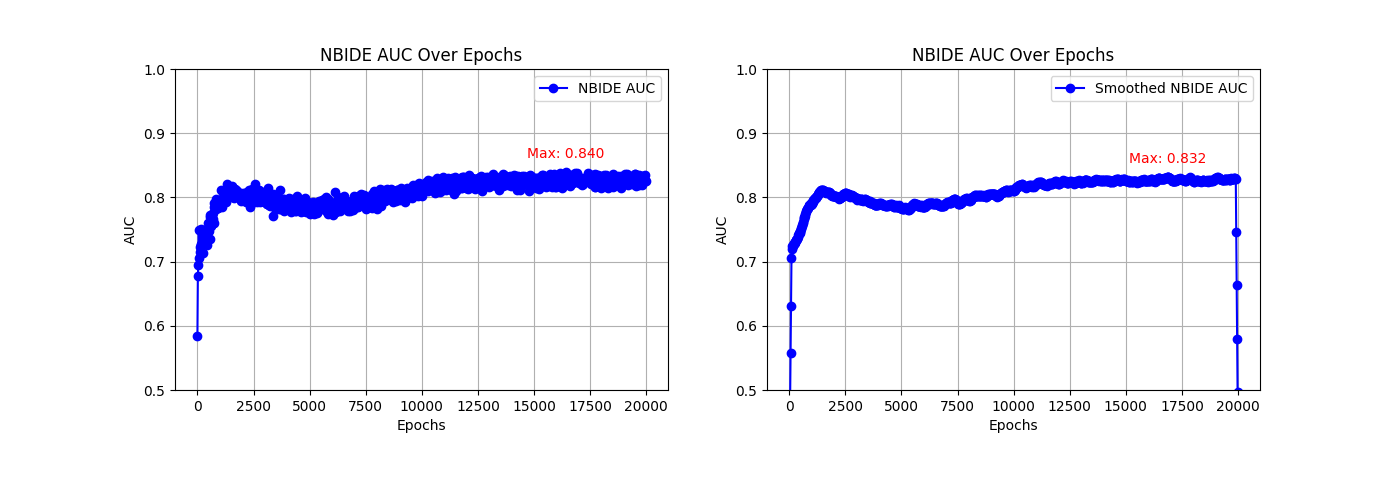}
        \caption*{(e) Width=8 model fifth run.}
    \end{minipage}

    \caption{Plots of the ROC AUC attained by the Width=8 model during 20,000 training epochs.}
    \label{fig:width_8_plots}
\end{figure}

\begin{figure}[htpb]
    \centering
    \hspace*{\fill} 
    
    \begin{minipage}[t]{0.45\textwidth} 
        \centering
        \includegraphics[width=\textwidth]{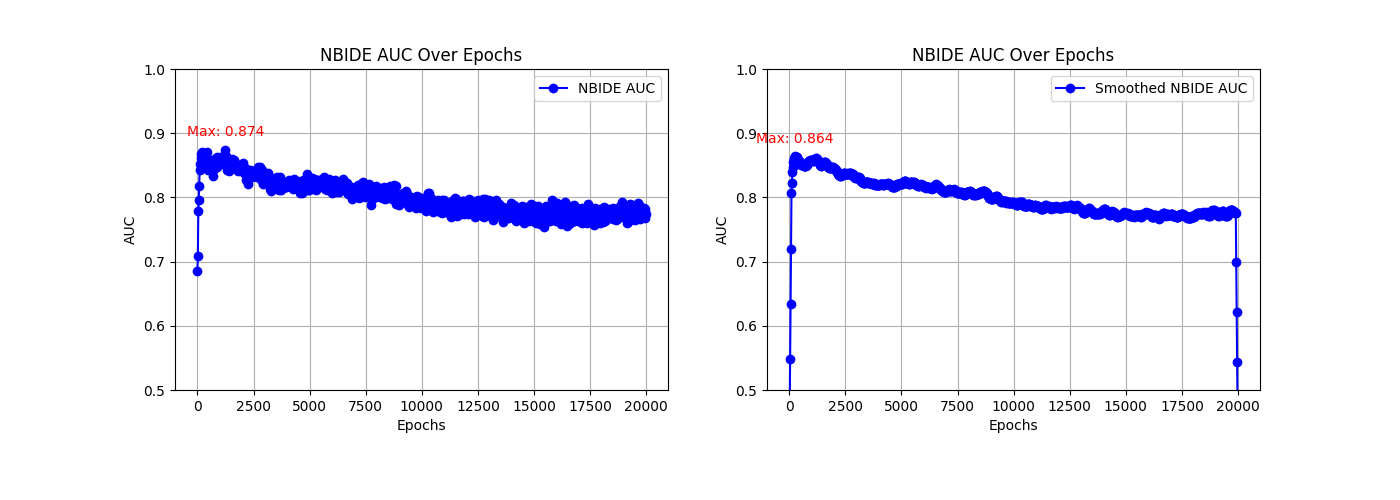}
        \caption*{(a) Width=32 model first run.}
        \includegraphics[width=\textwidth]{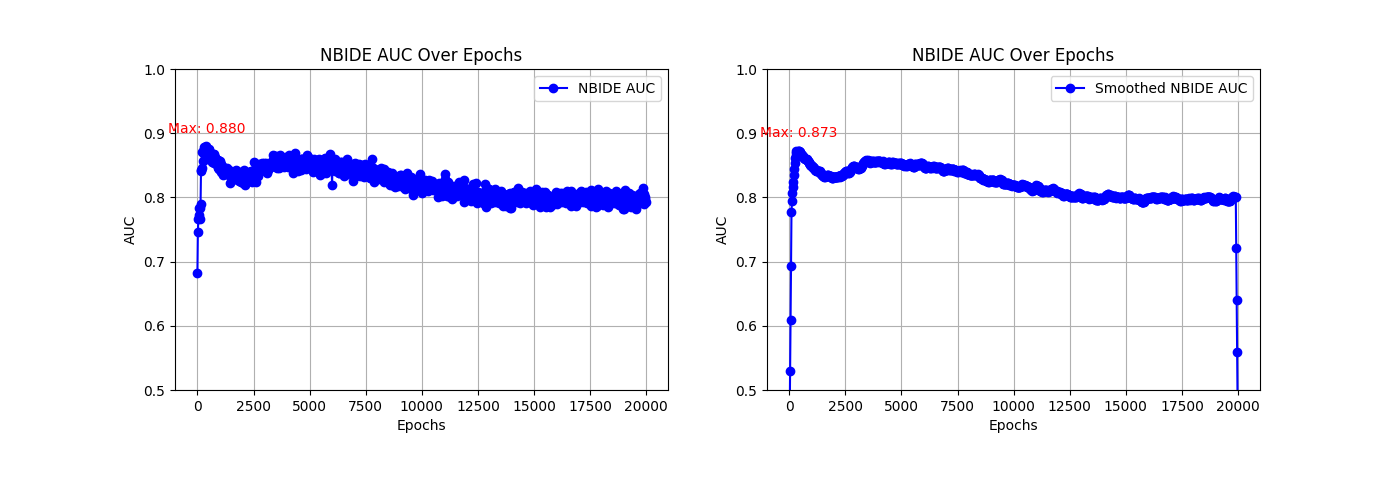}
        \caption*{(c) Width=32 model third run.}
    \end{minipage}
    \hfill
    \begin{minipage}[t]{0.45\textwidth} 
        \centering
        \includegraphics[width=\textwidth]{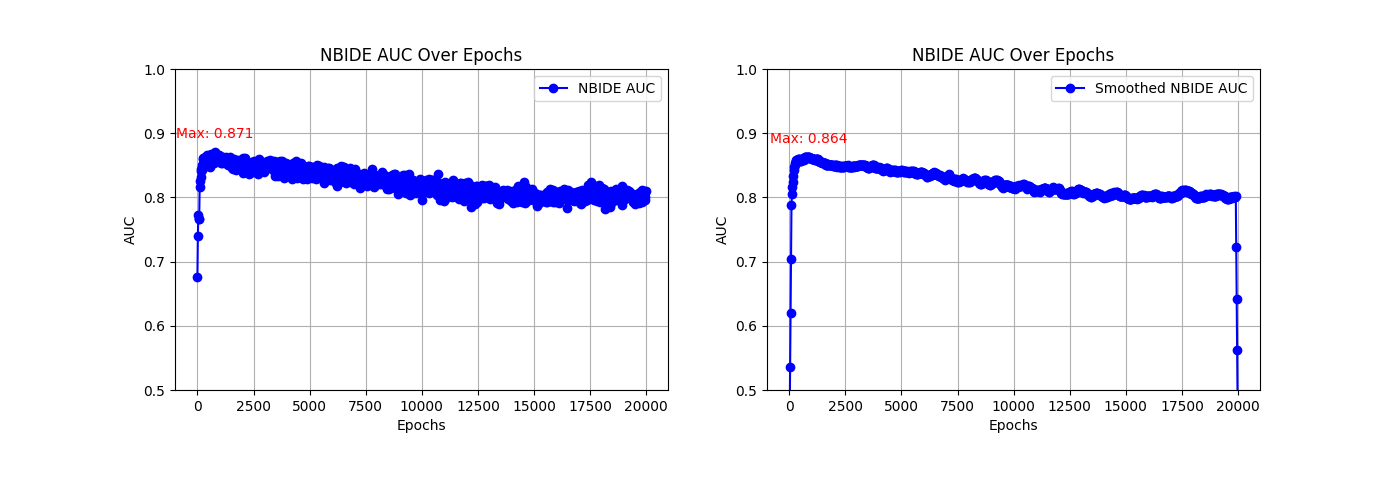}
        \caption*{(b) Width=32 model second run.}
        \includegraphics[width=\textwidth]{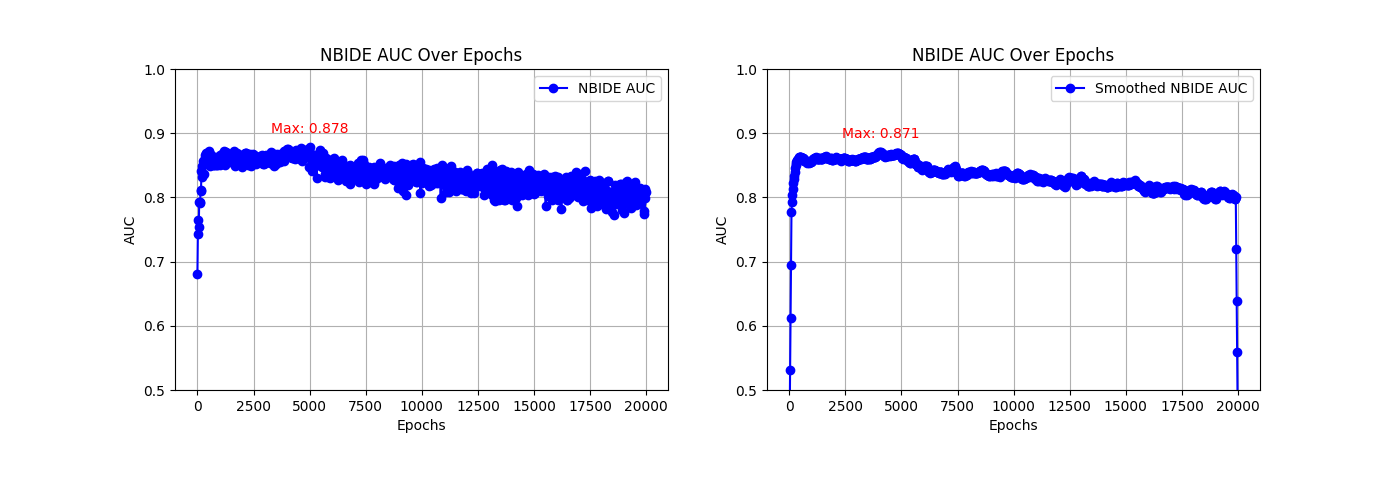}
        \caption*{(d) Width=32 model fourth run.}
    \end{minipage}
    
    \hspace*{\fill} 

    \vspace{.8em} 
    \begin{minipage}[t]{0.45\textwidth}
        \centering
        \includegraphics[width=\textwidth]{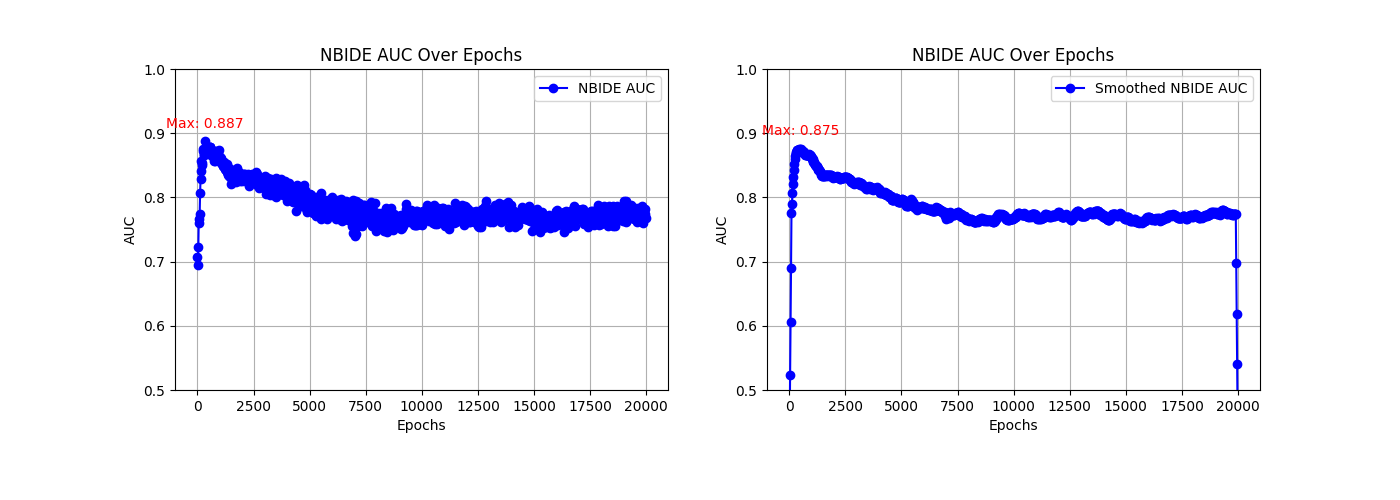}
        \caption*{(e) Width=32 model fifth run.}
    \end{minipage}

    \caption{Plots of the ROC AUC attained by the Width=32 model during 20,000 training epochs.}
    \label{fig:width_32_plots}
\end{figure}

\begin{figure}[htpb]
    \centering
    \hspace*{\fill} 
    
    \begin{minipage}[t]{0.45\textwidth} 
        \centering
        \includegraphics[width=\textwidth]{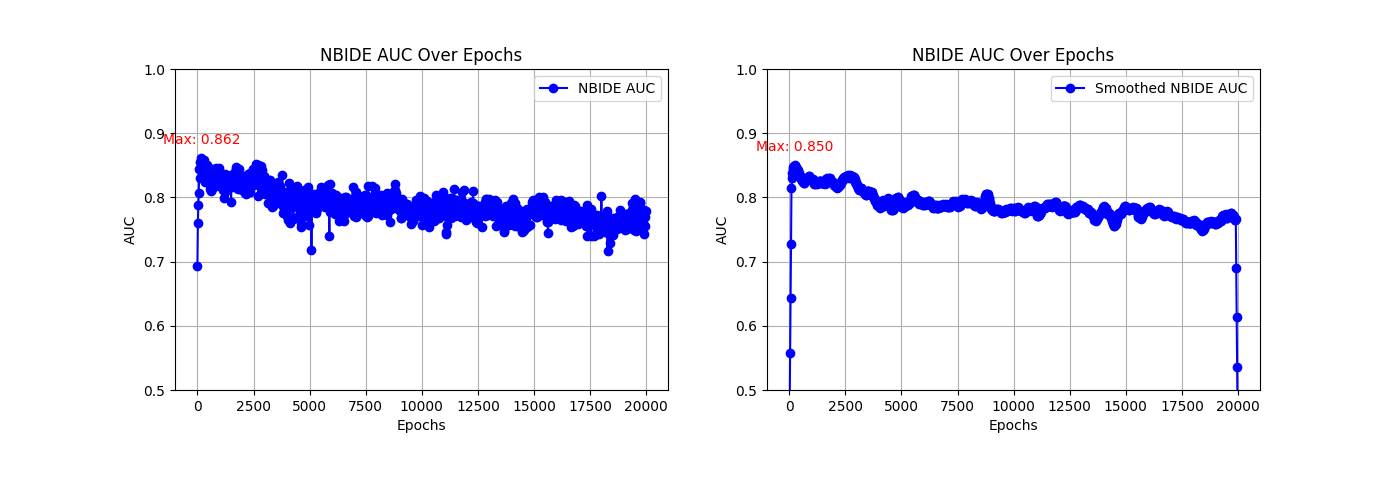}
        \caption*{(a) Width=64 model first run.}
        \includegraphics[width=\textwidth]{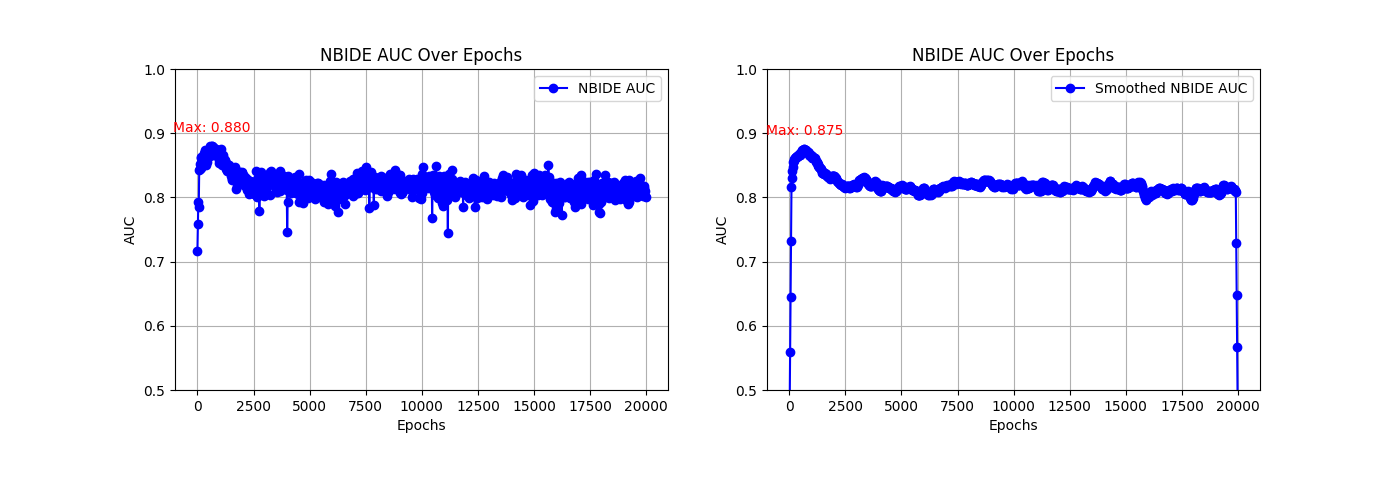}
        \caption*{(c) Width=64 model third run.}
    \end{minipage}
    \hfill
    \begin{minipage}[t]{0.45\textwidth} 
        \centering
        \includegraphics[width=\textwidth]{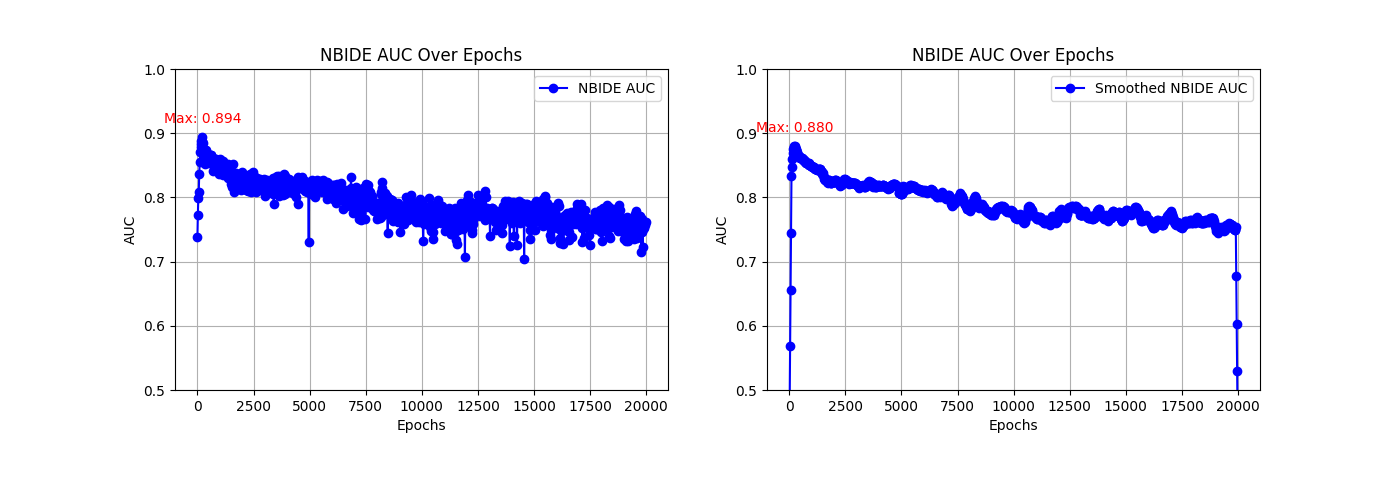}
        \caption*{(b) Width=64 model second run.}
        \includegraphics[width=\textwidth]{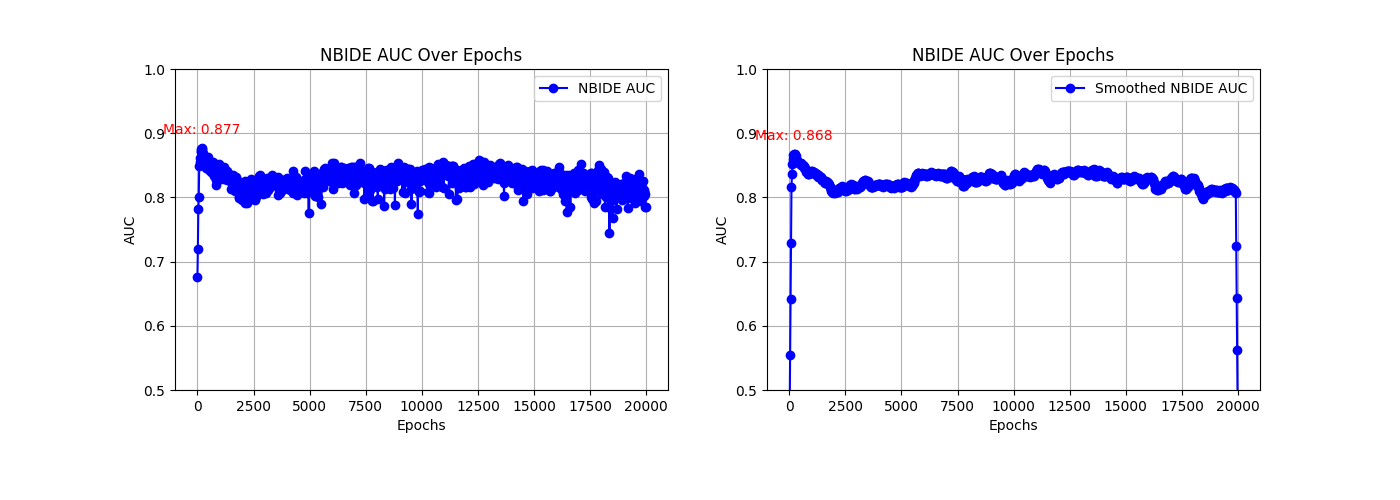}
        \caption*{(d) Width=64 model fourth run.}
    \end{minipage}
    
    \hspace*{\fill} 

    \vspace{.8em} 
    \begin{minipage}[t]{0.45\textwidth}
        \centering
        \includegraphics[width=\textwidth]{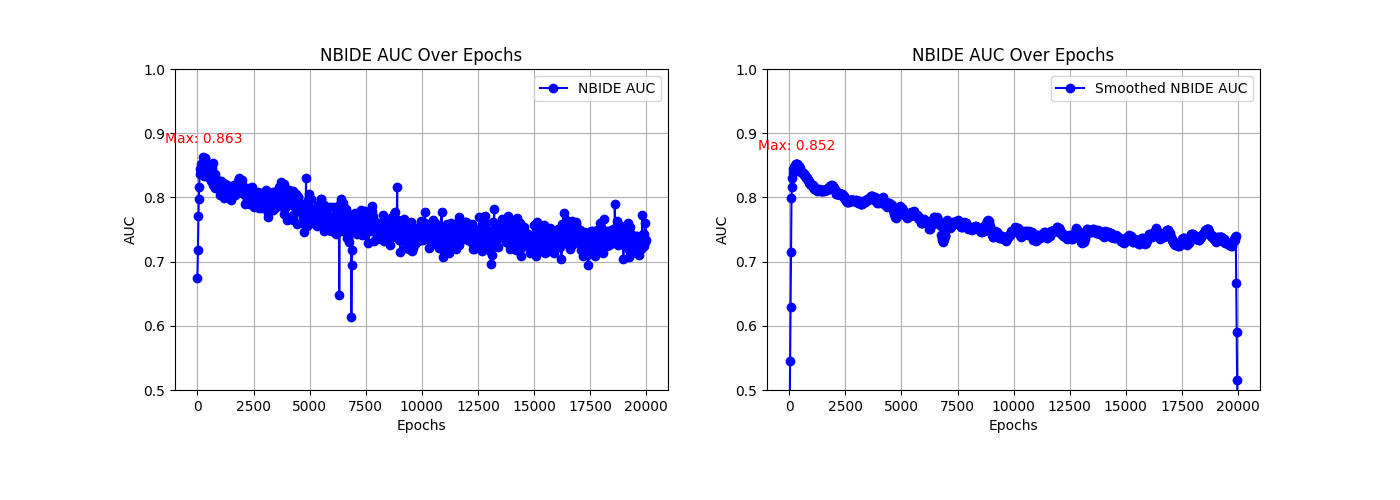}
        \caption*{(e) Width=64 model fifth run.}
    \end{minipage}

    \caption{Plots of the ROC AUC attained by the Width=64 model during 20,000 training epochs.}
    \label{fig:width_64_plots}
\end{figure}

\end{appendices}

\end{document}